\documentclass[twocolumn]{fairmeta}
\newcommand{\ours}{DexWM\xspace}
\usepackage{booktabs}
\usepackage{multirow}
\usepackage[table]{xcolor}
\usepackage{caption}
\usepackage{wrapfig}
\usepackage{xspace}
\usepackage{cuted}
\usepackage{graphicx}
\usepackage{amsmath}
\usepackage{amssymb}
\usepackage{float} 

\title{World Models for Learning Dexterous Hand-Object Interactions from Human Videos}

\author[1,2, *]{Raktim Gautam Goswami}
\author[1]{Amir Bar}
\author[1]{David Fan}
\author[1]{Tsung-Yen Yang}
\author[1,2]{Gaoyue Zhou}
\author[2]{Prashanth Krishnamurthy}
\author[1]{Michael Rabbat}
\author[2]{Farshad Khorrami}
\author[1,2]{Yann LeCun}

\affiliation[1]{FAIR at Meta}
\affiliation[2]{New York University}
\contribution[*]{Work done during internship at Meta}

\abstract{
\vspace*{-0.3cm}
Modeling dexterous hand–object interactions is challenging as it requires understanding how subtle finger motions influence the environment through contact with objects. While recent world models address interaction modeling, they typically rely on coarse action spaces that fail to capture fine-grained dexterity. We, therefore, introduce \ours, a Dexterous Interaction World Model that predicts future latent states of the environment conditioned on past states and dexterous actions. To overcome the scarcity of finely annotated dexterous datasets, \ours represents actions using finger keypoints extracted from egocentric videos, enabling training on over $900$ hours of human and non-dexterous robot data. Further, to accurately model dexterity, we find that predicting visual features alone is insufficient; therefore, we incorporate an auxiliary hand consistency loss that enforces accurate hand configurations. \ours outperforms prior world models conditioned on text, navigation, or full-body actions in future-state prediction and demonstrates strong zero-shot transfer to unseen skills on a Franka Panda arm with an Allegro gripper, surpassing Diffusion Policy by over 50\% on average across grasping, placing, and reaching tasks.
\vspace*{-0.5cm}
}

\metadata[Project Page]{\url{https://raktimgg.github.io/dexwm/}}

\begin{document}

\maketitle

\begin{strip}
\centering
\vspace*{-0.5cm}
\includegraphics[trim={0 0 0 0},clip, width=\textwidth]{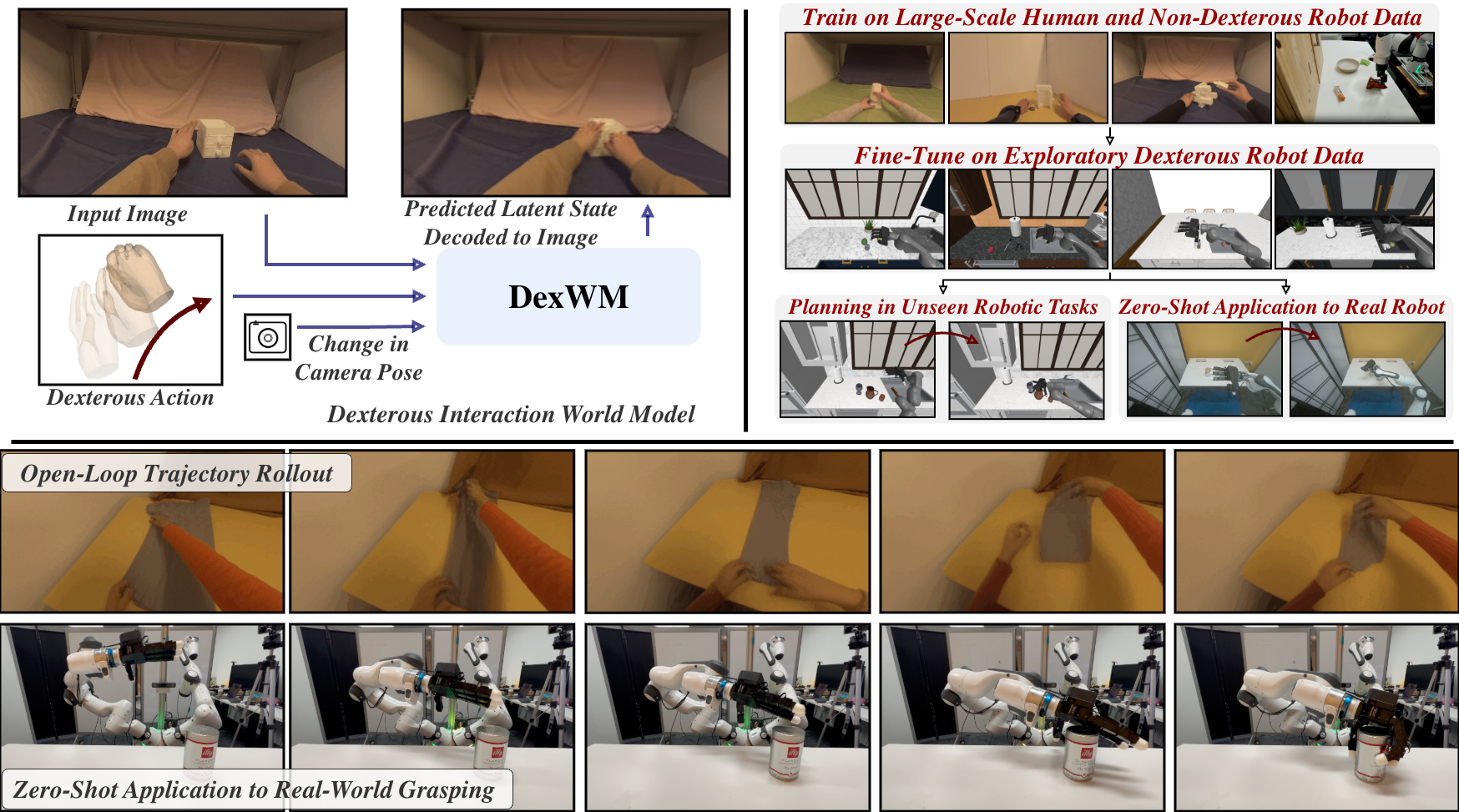}
\captionsetup{type=figure}
\vspace*{-0.5cm}
\caption{We introduce \textbf{\ours}, a Dexterous Interaction World Model which predicts future latent states of the environment from past states and dexterous actions. Trained on large-scale human and non-dexterous robot videos, \ours learns to simulate complex trajectories. Further, with minimal fine-tuning on a small exploratory robot simulation dataset, \ours enables planning for novel reaching, grasping, and placing tasks in simulation, and transfers zero-shot to real-world robot grasping.}
\label{fig:teaser}    
\end{strip}

\section{Introduction}
As embodied agents become increasingly integrated into daily lives, dexterous manipulation emerges as a critical capability for achieving human-like interaction with the physical world. Everyday tasks like cooking, as well as high-stakes applications like surgery, demand a level of dexterity that is infeasible with commonly used parallel-jaw grippers. 
Modeled after the human hand, dexterous grippers enable complex tool use, fine-grained movements, and in-hand manipulation~\cite{dexterous_rl_review, han1998dextrous, in_hand_manipulation, morgan2022complex}. 
Recent advances in deep learning have produced vision-based manipulation policies~\cite{diff_policy,lee2024behaviorgenerationlatentactions, haldar2024bakuefficienttransformermultitask, kim2024openvlaopensourcevisionlanguageactionmodel,embodimentcollaboration2025openxembodimentroboticlearning}, yet these methods often struggle to generalize to unseen tasks and to plan robustly in physical environments~\cite{awda, osvi-wm, zhou2023trainofflinetestonline}. Successful execution requires models that reason about how actions affect the environment, such as recognizing that opening the hands when holding an object will cause it to drop.

World models can learn environmental dynamics from observation and action~\cite{lecun2022path}, and thus offer a promising solution. 
Early work on learned world models~\cite{transformer_wm, dino-wm, gaia} has primarily focused on small-scale tasks with constrained environments and limited action spaces. More recent efforts handle complex actions, such as text~\cite{agarwal2025cosmos}, navigation~\cite{bar2025navigation}, and whole-body motion~\cite{peva}. 
However, the action representations in these methods are typically too coarse to capture the fine-grained structure required for modeling dexterous dynamics. Similar to prior work in world models~\cite{bar2025navigation, dino-wm, vjepa2, peva}, we define ``dynamics'' as the modeling of how the agent moves in response to actions, how the environment reacts, and how both appear from the camera's viewpoint.
An important application of such dexterous world models is robotic manipulation. However, learning them directly from robot data is challenging due to the lack of large-scale datasets with dexterous grippers.

To address these challenges, we propose \ours (\Cref{fig:teaser}), a latent space world model that learns from human data to predict future latent states based on past states and dexterous hand actions. Inspired by recent work~\cite{hop, maple} that leverages human training data, we pre-train \ours on EgoDex~\cite{egodex}, a large-scale egocentric human interaction dataset, and further incorporate DROID~\cite{droid} sequences, consisting of non-dexterous robot manipulation, to learn cross-embodiment dynamics. 
\ours's actions are represented as differences in 3D hand keypoints and camera poses, capturing detailed hand configurations and enabling the model to learn how hand posture changes affect the environment. 

We find that accurately simulating hand locations using the next latent state prediction objective alone is difficult. Therefore, we train \ours to jointly optimize the future environment state and the hand configuration, providing a richer learning signal for dexterity. \ours outperforms prior world models~\cite{bar2025navigation,peva,agarwal2025cosmos} in open-loop trajectory simulation, better captures fine-grained hand dynamics, and demonstrates reliable controllability under arbitrary action sequences.

Furthermore, \ours enables strong zero-shot transfer to robot manipulation tasks of grasping, placing, and reaching by optimizing actions at test time within a Model Predictive Control (MPC) framework. 
Notably, when deployed on a real-world Franka Panda robot with the Allegro grippers, \ours achieves an 83\% success rate in object grasping. To bridge the gap between training data and robot embodiment, we fine-tune the model on about four hours of exploratory, non-task-specific dexterous data from the RoboCasa simulation suite~\cite{robocasa}. Unlike existing behavior cloning methods~\cite{hop,maple,diff_policy} that predict actions or waypoints from observation, \ours is used as a state-transition model within an MPC optimization framework for planning waypoint trajectories, which are executed via low-level controllers, thus offering greater robustness.

In summary, we introduce \ours, a latent-space world model for learning dexterous hand–object interaction dynamics directly from human videos. To support fine-grained dexterity, we incorporate an auxiliary hand-consistency loss. Furthermore, \ours demonstrates that world models trained on human data can scale effectively and transfer to zero-shot robotic manipulation tasks such as reaching, grasping, and placing.

\section{Related Work}

Recent environment modeling approaches are largely built on Diffusion and Flow Matching objectives, which learn to generate videos by denoising or integrating learned velocity fields over time~\cite{sohl2015deep,ho2020denoising,song2020score,liu2022flow,esser2024scaling,peebles2023scalable}. Combined with large-scale text–video training, these models have improved to the point where they can simulate highly realistic videos from textual prompts~\cite{hong2022cogvideo,blattmann2023align,yang2024cogvideox,opensora,videoworldsimulators2024}. For interactive and streaming applications~\cite{qiu2023freenoise,chen2023seine,wang2023gen,zhang2024avid,chen2024streaming}, generating video frames autoregressively is particularly important, but introduces error accumulation over time~\cite{yan2021videogpt,liang2022nuwa,ge2022long,weng2024art,gao2024vid}. Diffusion Forcing~\cite{chen2024diffusion} addresses this issue by combining next-token prediction with full-sequence diffusion and injecting noise into previously generated context frames, while other autoregressive video diffusion approaches mitigate drift through multistep prediction and refinement~\cite{xie2024progressive,henschel2024streamingt2v,kim2024fifo}. In this work, we focus primarily on building such action-conditioned predictive models for dexterous hand-object interactions.

Action-conditioned predictive models, often referred to as ``World Models'', have recently been used to simulate computer game engines~\cite{yu2025gamefactory,hunyuanworld,bruce2024genie}. For example, DIAMOND~\cite{diamond} aims to simulate Counter-Strike, while Dreamer~\cite{hafner2025training} has been applied to MineCraft. Other approaches focus on more visually realistic settings with continuous action spaces, including navigation~\cite{hu2023gaia1generativeworldmodel,bar2025navigation,genex} and full-body control~\cite{peva}. In contrast, modeling dexterous dynamics requires precise, fine-grained control over hand–object interactions, and remains more challenging.

World models have also become increasingly influential in robotics. While commonly used to train reinforcement learning agents~\cite{dreamer, rssm, td-mpc2, transdreamer}, they have also proven effective for model-based policy optimization~\cite{dino-wm, vjepa2}. 
For dexterous dynamics modeling, prior works in world models have explored cross-embodiment dynamics via particle-based representations~\cite{he2025scaling, honglearning}, goal-conditioned manipulation through articulated object modeling (DexSim2Real$^2$~\cite{jiang2409dexsim2real2}), and scene-action-conditioned video diffusion for simulating dexterous behavior (DWM~\cite{kim2026dwm}).


In addition to modeling dexterous dynamics, we show that \ours can be used in a Model Predictive Control (MPC) framework for zero-shot dexterous manipulation tasks such as grasping.
Dexterous manipulation, in general, remains a longstanding challenge in robotics. Early approaches relied on hand-crafted gaiting methods for controlling grippers~\cite{dex_early_work1, dex_early_work2, dex_early_work3}, but the high degrees of freedom and complex contact dynamics of such grippers has led to a shift toward learning-based methods~\cite{yu2022dexterous, an2025dexterous}. Sim2real techniques, in particular, have enabled robots to perform complex tasks such as solving a Rubik’s cube~\cite{akkaya2019solving}, in-hand manipulation~\cite{dextreme, visual_dexterity, general_in_hand}, and functional grasping~\cite{functional_grasping}.
Much of this progress is driven by improved access to training data. Although datasets with dexterous manipulation remain scarce, the availability of large-scale egocentric human video datasets~\cite{egodex, hot3d, ego4d, 100days} and efficient hand annotation tools~\cite{mano, hamer, dynhamr} has enabled recent research~\cite{yang2015robot, bahl2023affordances, chen2025vividex, radosavovic2023robot, kareer2025egomimic, qin2022dexmv, shaw2023videodex, egovla, groot} to leverage human demonstrations for learning dexterous manipulation. HOP~\cite{hop}, for example, retargets human videos to robot simulation for sensorimotor learning, while MAPLE~\cite{maple} pre-trains encoders with dexterous priors. 
Similarly, we train \ours on egocentric human videos~\cite{egodex} and sequences of non-dexterous robot data~\cite{droid} to learn dexterous manipulation dynamics.

\section{Proposed Methodology}
\label{sec:methods}
\begin{figure}
    \centering
    \includegraphics[width=\linewidth]{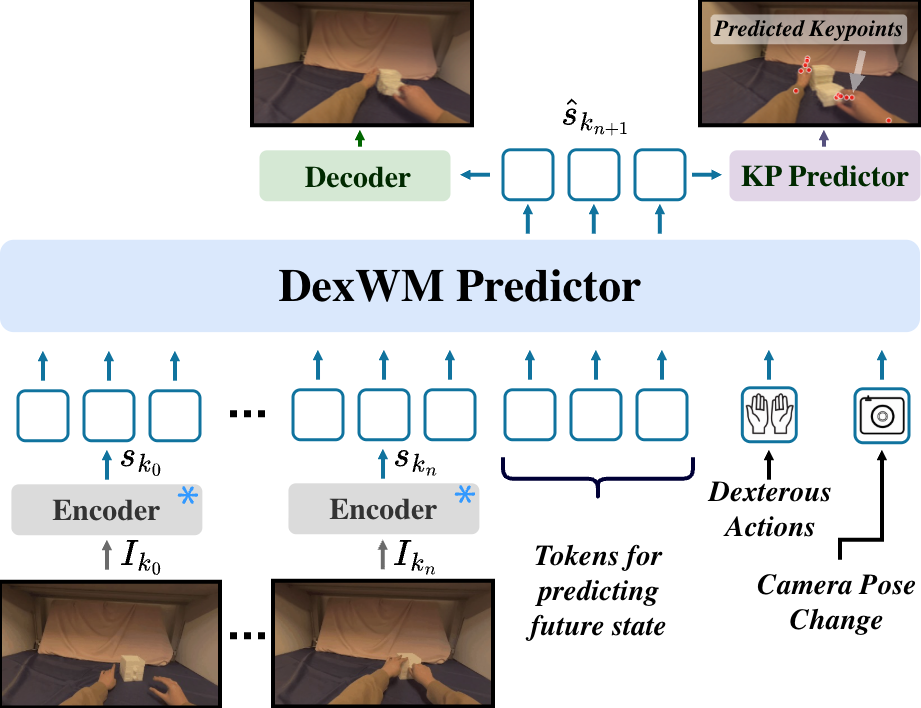}
    \caption{\textbf{\ours Architecture.} Images are encoded into latent states using a frozen DINOv2~\cite{DINOv2} encoder. The \ours predictor takes these states, hand actions, and camera motions to predict the next state, which can then be decoded into reconstructed images and hand keypoints.}
    \label{fig:flow}
    \vspace*{-0.5cm}
\end{figure}

Our goal is to build a world model that predicts how dexterous hand–object interactions unfold: how the hand moves, how objects respond, and how both appear from an egocentric viewpoint. A schematic illustration of the model is shown in~\Cref{fig:flow}.
In what follows, we first describe the problem formulation, then present the State and Action representations (Sec~\ref{sec:states_and_actions}), the Predictor (Sec~\ref{sec:predictor}), and the Loss (Sec~\ref{sec:training_loss}). Finally, we explain how to use a trained \ours for planning manipulation in Sec~\ref{sec:planning}.

\vspace{1.3mm}
\noindent\textbf{Problem Formulation.}
\label{sec:problem_formulation}
Let $s_{k_i} \in \mathcal{S}$ denote the environment's (latent) state at timestep $k_i$, for index $i$, which encodes the configuration of the dexterous hand and the geometry of the surrounding scene. The agent issues a dexterous action $a_{k_1 \to k_2} \in \mathcal{A},$ specifying the hand control executed between timesteps $k_1$ and $k_2$.

Since the latent state $s_{k_i}$ is not directly observed, the agent receives an egocentric RGB image $I_{k_i} \in \mathbb{R}^{H \times W \times 3},$ captured by a human or robot equipped with a dexterous hand. We introduce an encoder $E_\phi : \mathbb{R}^{H \times W \times 3} \to \mathcal{S},$ which maps the observation $I_{k_i}$ to a latent state representation $s_{k_i} = E_\phi(I_{k_i})$.

To approximate the environment dynamics, we define the predictor $f_\theta : \mathcal{S} \times \mathcal{A} \to \mathcal{S},$
as a mapping from a current latent state and an action to a predicted future state.\footnote{$f_\theta$ can take multiple temporal states as input; \Cref{eq:pf} shows a single for simplicity.} Formally, our objective is to learn 
\begin{align}
    \hat{s}_{k_2} = f_\theta(s_{k_1}, a_{k_1 \to k_2})
        = f_\theta(E_\phi(I_{k_1}), a_{k_1 \to k_2}),
    \label{eq:pf}
\end{align}
where $\hat{s}_{k_2}$ is the predicted environment state after executing $a_{k_1 \to k_2}$.

\subsection{State and Action Representation}
\label{sec:states_and_actions}
Next, we describe the states and actions used in~\ours.

\vspace{1.3mm}
\noindent\textbf{Latent States.} Pixel-level details are often unnecessary for modeling the underlying system dynamics. For example, in object manipulation, the agent is typically more concerned with the object's shape and structure than its color. 
To address this, we employ the DINOv2~\cite{DINOv2} image encoder ($E_\phi$) to transform pixel data into a latent embedding space following~\cite{dino-wm,vjepa2,osvi-wm}. Specifically, we utilize patch-level features as the latent state such that $s_{k_i} \in \mathbb{R}^{\mathcal{P} \times d}$, where $\mathcal{P}$ is the number of image patches and $d$ is the feature dimension per patch. DINOv2 features are semantically rich and have demonstrated strong generalization across diverse environments and scenarios~\cite{DINOv2,dino-wm}.

\vspace{1.3mm}
\noindent\textbf{Action Representation.} A key challenge is how to represent actions in a dexterous dynamics model. Prior work has modeled wrist position~\cite{peva} or used high-level text~\cite{agarwal2025cosmos}, but these representations are often too coarse for dexterous skills. Instead, we seek an action representation that precisely captures the change of the agent’s hands, as well as how they move while performing a task.

\begin{figure}[t]
    \centering
    \begin{subfigure}{\linewidth}
        \centering
        \includegraphics[trim={0 30 0 0},clip,width=\linewidth]{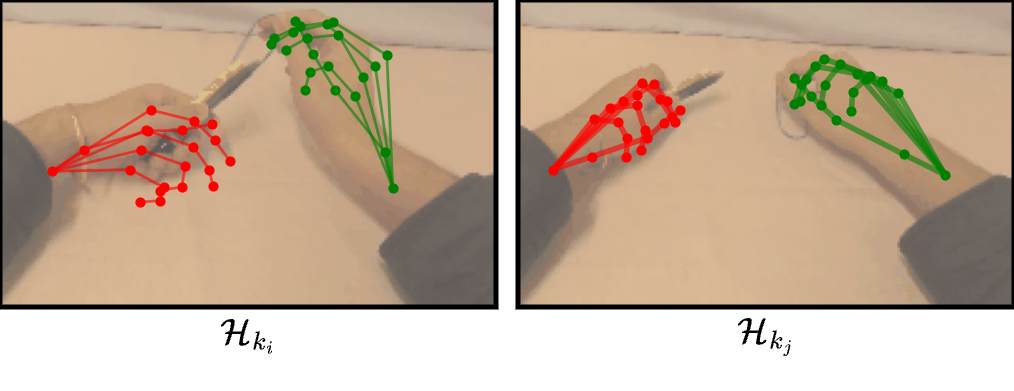}
        \caption{}
        \vspace*{-0.3cm}
        \label{fig:keypoints}
    \end{subfigure}

    \vspace{0.2cm}

    \begin{subfigure}{\linewidth}
        \centering
        \includegraphics[trim={0 0 0 0},clip,width=\linewidth]{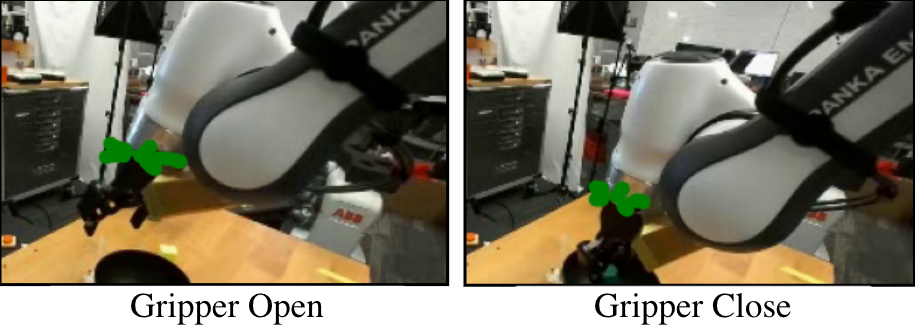}
        \caption{}
        \vspace*{-0.3cm}
        \label{fig:droid_hands}
    \end{subfigure}

    \caption{\textbf{(a)} \textbf{Action Representation.} Hand actions are represented as differences in 3D keypoints between frames (e.g., $H_{k_j} - H_{k_i}$), providing a unified representation of dexterous actions in \ours. This is supplemented with camera motion to capture the agent’s movement. \textbf{(b)} For the DROID~\cite{droid} dataset, parallel-jaw grippers are approximated as dexterous hands using dummy keypoints placed on concentric circles at the end-effector, with radii varying by the gripper’s open/close state, mimicking finger spread.}
    \vspace*{-0.5cm}
\end{figure}

We start by defining the action vector between states $s_{k_1}$ and $s_{k_2}$ as the difference in keypoint positions between timesteps ${k_1}$ and ${k_2}$. 
Following the MANO~\cite{mano} parameterization, each hand is represented by $21$ keypoints (Fig.~\ref{fig:keypoints}), with coordinates $\mathcal{H}_{k_i}^L, \mathcal{H}_{k_i}^R \in \mathbb{R}^{21 \times 3}$ for the left and right hands, respectively, at state $s_{k_i}$. Specifically, $\mathcal{H}_{k_i} = \{\mathcal{H}_{k_i}^L, \mathcal{H}_{k_i}^R\}$ encodes the hand configurations. 

However, simply using hands information to represent actions does not account for how the agent moves. To address this, we apply two modifications. First, instead of representing $\mathcal{H}_{k_2}$ in the camera frame at ${k_2}$, we use the known rigid transformation $T_{k_2}^{k_1}$ to express all keypoints in the same coordinate frame ${k_1}$. Second, to inform the world model about camera pose changes, we append the change in camera translation $\delta t_{k_1\to k_2} \in \mathbb{R}^3$ and orientation $\delta q_{k_1\to k_2} \in \mathbb{R}^3$ (as Euler angles) to the action vector. The final action vector is defined as
\begin{align}
    a&_{{k_1} \to {k_2}} = \nonumber
    \\ &\left[ (\mathcal{H}_{k_2} - \mathcal{H}_{k_1})^T, \ \delta t_{k_1\to k_2}^T, \ \delta q_{k_1\to k_2}^T \right]^T \in \mathbb{R}^{44 \times 3}.
\end{align}

Most egocentric human video datasets~\cite{egodex, hot3d, holoassist} already include hand keypoint annotations. For the DROID~\cite{droid} dataset, parallel-jaw grippers are approximated as dexterous hands using dummy keypoints at the end-effector (\Cref{fig:droid_hands}). For real robot and RoboCasa~\cite{robocasa} simulation data, keypoints are computed via forward kinematics from known joint angles. To map the robot’s 4-finger Allegro hand to DexWM’s 5-finger action space, the last Allegro finger (equivalent to the human ring finger) keypoints are also used to represent the pinky finger.

\subsection{Predictor}
\label{sec:predictor}
With the states and actions of the world model defined, we now describe the predictor model $f_\theta$ used in \ours. The predictor model takes as input a history of observed environment states $s_{k_0}, \dots, s_{k_n}$ and current action $a_{k_n \to k_{n+1}}$, and predicts the next state $\hat{s}_{k_{n+1}}$. Formally,
\begin{align}
    \hat{s}_{k_{n+1}} = f_\theta(s_{k_0}, \dots, s_{k_n}, a_{k_n \to k_{n+1}})
\end{align}
where $\theta$ denotes the trainable parameters of the network.

Unlike previous works~\cite{bar2025navigation,peva}, we assume that the environment is deterministic, which leads to faster inference time. Also, while the timesteps $k_0, \dots, k_n$ can be uniformly spaced, we find that training in a non-fixed frequency by randomly skipping states improves generalization.
 
Our predictor architecture is based on Conditional Diffusion Transformers (CDiT)~\cite{bar2025navigation}. CDiT provides strong action conditioning via AdaLN~\cite{peebles2023scalable} layers, where the flattened action vector of $44 \times 3=132$ dimensions is projected as conditioning input to each transformer block. Unlike diffusion-based CDiT methods~\cite{bar2025navigation,peva}, we directly regress future latent states using DINOv2 features for faster inference without iterative denoising. To use CDiT with DINOv2, we define tokens for predicting the future state $\hat{s}_{k_{n+1}}$ (\Cref{fig:flow}), initialized from $s_{k_n}$.

\vspace{1.3mm}
\noindent\textbf{Multistep Prediction.} For inference and planning in robotic tasks, we perform multistep prediction by autoregressively feeding the predicted state $\hat{s}_{k_{n+1}}$ and the subsequent action $a_{k_{n+1} \to k_{n+2}}$ back into $f_\theta$ to generate $\hat{s}_{k_{n+2}}$, continuing this process for future timesteps.

\subsection{Hand Consistency Training Loss}
\label{sec:training_loss}
As the primary goal of \ours is to predict future latent states of the environment, we use a mean squared error loss on the predicted embeddings:
\begin{align}
    \mathcal{L}_{\text{state}} = \frac{1}{\mathcal{P} \times d} \sum_{p=1}^{\mathcal{P}} \| s_{k_{n+1}}(p) - \hat{s}_{k_{n+1}}(p) \|_2^2
\end{align}
where $p$ indexes the image patches and $s_{k_{n+1}}$ is the DINOv2 embedding of the ground truth image.

However, we observe that relying solely on $\mathcal{L}_{\text{state}}$ is insufficient for capturing the fine-grained details necessary for modeling dexterous dynamics, as the hands occupy only a small region of the image. To address this, we incorporate an additional loss term that encourages the model to capture local information. Specifically, we use a transformer-based network $g_\theta$ to predict heatmaps of the fingertip and wrist locations in $\mathcal{H}_{k_{n+1}}$, denoted as $\hat{V}_{k_{n+1}} \in \mathbb{R}^{12 \times H \times W}$. This ensures that the predicted state $\hat{s}_{k_{n+1}}$ is informative enough to recover keypoint positions.
This hand consistency (HC) loss is defined as:
\begin{align}
    \mathcal{L}_{\text{HC}} = \frac{1}{12\times H\times W} \| V_{k_{n+1}} - \hat{V}_{k_{n+1}} \|_2^2
\end{align}
where $V_{k_{n+1}}$ are the ground truth heatmaps.
During training, the image encoder is kept frozen, while the rest of the model is optimized using $\mathcal{L} = \mathcal{L}_{\text{state}} + \lambda \mathcal{L}_{\text{HC}}$ with $\lambda = 100$ yielding the best results in our experiments.

\subsection{Robot Task Planning}
\label{sec:planning}
In addition to modeling dexterous interaction dynamics, \ours enables zero-shot transfer to robotic manipulation tasks. Specifically, \ours is used as a state transition model to plan waypoint trajectories for downstream control.

\begin{figure}
    \centering
    \includegraphics[width=.95\linewidth]{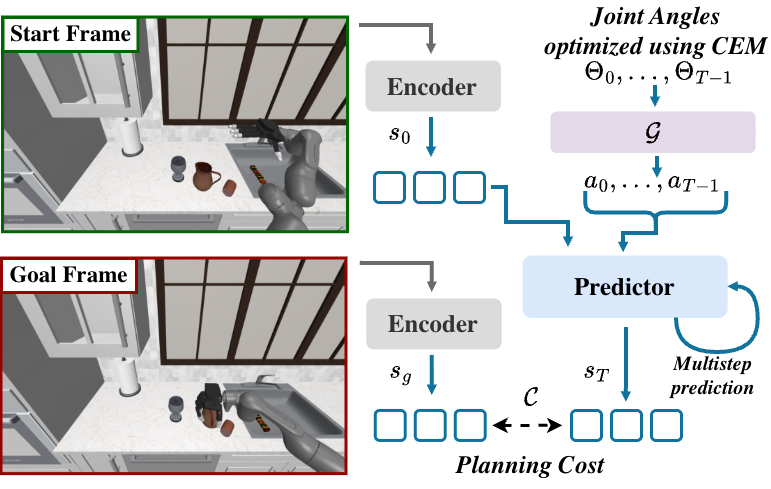}
    \caption{\textbf{Goal-Conditioned Planning with \ours.} The joint angles $\Theta_0, \dots, \Theta_{T-1}$ are optimized using CEM to get the optimal actions $a_0, \dots, a_{T-1}$ to drive the system to the goal.}
    \label{fig:planning}
    \vspace*{-0.5cm}
\end{figure}

\begin{table*}[t!]
\centering
\footnotesize
\setlength{\tabcolsep}{6pt} 
\begin{tabular}{ccccccccc}
\toprule
  & \multicolumn{4}{c}{EgoDex} & \multicolumn{4}{c}{RoboCasa} \\
  \cmidrule(lr){2-5} \cmidrule(lr){6-9}
 Dataset & \multicolumn{2}{c}{Embedding L2 Error$\downarrow$} & \multicolumn{2}{c}{PCK@20$\uparrow$} 
      & \multicolumn{2}{c}{Embedding L2 Error$\downarrow$} & \multicolumn{2}{c}{PCK@20$\uparrow$} \\
\cmidrule(lr){2-3} \cmidrule(lr){4-5} \cmidrule(lr){6-7} \cmidrule(lr){8-9}
 & At 4s & Avg & At 4s & Avg & At 4s & Avg & At 4s & Avg \\
\midrule
 EgoDex &  0.67&  0.51&  \cellcolor{gray!20}\textbf{60}&  68&  1.03&  0.79&  3&  13\\
 DROID &  1.39&  1.06&  1&  2&  1.3&  0.96&  2&  12\\
 EgoDex+DROID &  \cellcolor{gray!20}\textbf{0.66}&  \cellcolor{gray!20}\textbf{0.5}&  \cellcolor{gray!20}\textbf{60}& \cellcolor{gray!20} \textbf{69}&  \cellcolor{gray!20}\textbf{0.79}&  \cellcolor{gray!20}\textbf{0.57}&  \cellcolor{gray!20}\textbf{7}&  \cellcolor{gray!20}\textbf{17}\\
\bottomrule
\end{tabular}
\caption{\textbf{\ours Benefits From Human Video Data.} Training \ours on EgoDex in addition to DROID contributes to downstream open-loop performance in RoboCasa, as measured by lower embedding loss and higher PCK@20.}
\label{tab:dataset_ablation}
\vspace*{-0.4cm}
\end{table*}

\vspace{1.3mm}
\noindent \textbf{Planning Optimization.}
We use a goal-conditioned planning setup with the learned model $f_\theta$. Given initial ($s_0$) and goal ($s_g$) states, we solve:
\begin{align}
    &\Theta_0^*, \dots, \Theta_{T-1}^* = \arg\min_{\Theta_0, \dots, \Theta_{T-1}} \; \mathcal{C}(s_T, s_g) \nonumber \\
    \text{s.t} \quad a_k &= \mathcal{G}(\Theta_k), \\
    \quad \hat{s}_{k+1} &= f_\theta(s_k,\dots,s_0, a_k), \quad k = 0, \dots, T-1
    \nonumber
\end{align}
where $\Theta_k$ represent the joint angles of the robot, $\mathcal{G}$ uses the forward kinematics of the robot to calculate $a_k$, $\mathcal{C}$ is the planning cost, and $T$ is the planning horizon (\Cref{fig:planning}). In this formulation, we assume uniformly sampled timesteps with states $\{s_0, s_1, \dots\}$ and actions $\{a_0, a_1, \dots\}$. We use the Cross-Entropy Method (CEM)~\cite{crossentropymethod} to optimize the joint angles. Implementation details of CEM, including the number of samples, planning horizon, iteration count, and computational cost, are provided in the appendix.

\vspace{1.3mm}
\noindent \textbf{Planning Cost.} We use the planning cost $\mathcal{C} = \mathcal{C}_{\text{state}} + \mu \mathcal{C}_{\text{kp}}$ ($\mu = 0.001$), with $\mathcal{C}_{\text{state}}$ being the $L_2$ distance between latent states $s_T$ and $s_g$, and $\mathcal{C}_{\text{kp}}$ the Euclidean distance between keypoint pixels locations corresponding to heatmaps $\hat{V}_T$ and $\hat{V}_g$ predicted from $s_T$ and $s_g$, respectively. Combining both cost terms improves planning performance over using $\mathcal{C}_{\text{state}}$ alone, indicating latent embeddings alone may be suboptimal for planning. For the grasping task, we also add a cost on end-effector orientation to maintain neutral poses for successful grasps.

\section{Experiments and Results}
In this section, we first analyze the contribution of \ours’s individual components. We then evaluate its performance in open-loop dexterous rollouts and show its ability to capture fine-grained hand dynamics and maintain reliable control under arbitrary atomic actions. Finally, we assess \ours in zero-shot trajectory planning across both simulated and real-world robotic settings.

\subsection{Datasets}
\label{sec:dataset}

For training and evaluation, we use EgoDex~\cite{egodex}, an egocentric human video dataset containing $829$ hours of 1080p footage with rich hand and pose annotations, and DROID~\cite{droid}, a diverse dataset of robot manipulation with parallel-jaw grippers. We also use about 4 hours of exploratory sequences of random arm motions collected in the RoboCasa~\cite{robocasa} simulation framework using a Franka Panda arm with Allegro gripper for fine-tuning the model for robotic tasks. For more details, see the appendix.

\subsection{Ablation Studies}
\label{sec:ablation}
\noindent\textbf{Human Video.} We evaluate the impact of training on human videos to downstream performance (see \Cref{tab:dataset_ablation}). We assess zero-shot open-loop rollout performance on \textit{Lift}, a dataset collected in RoboCasa (see appendix), and on EgoDex. We report embedding $L_2$ loss and PCK@20 over predicted keypoints $\hat{V}$ (\Cref{sec:training_loss}) at $4$ seconds and on average. As shown in \Cref{tab:dataset_ablation}, adding EgoDex on top of DROID significantly boosts performance on RoboCasa, while preserving performance on EgoDex. This indicates that adding human video contributes to downstream performance across different embodiments.

\begin{figure}[t]
    \centering
    \includegraphics[trim={0 10 0 0},clip, width=\linewidth]{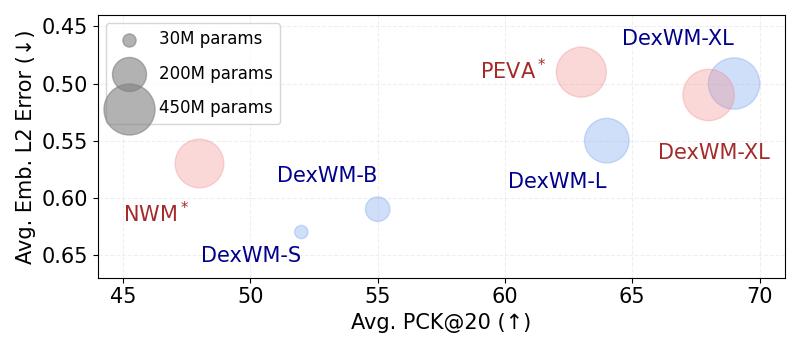}
    \caption{\textbf{Scaling With Predictor Size.} PCK@20 and Embedding L2 error improve with larger \ours models. Blue circles denote EgoDex+DROID training; red circles denote EgoDex-only training.}
    \label{fig:model_size}
\end{figure}

\begin{figure}[t]
    \centering
    \includegraphics[trim={0 0 0 0},clip, width=\linewidth]{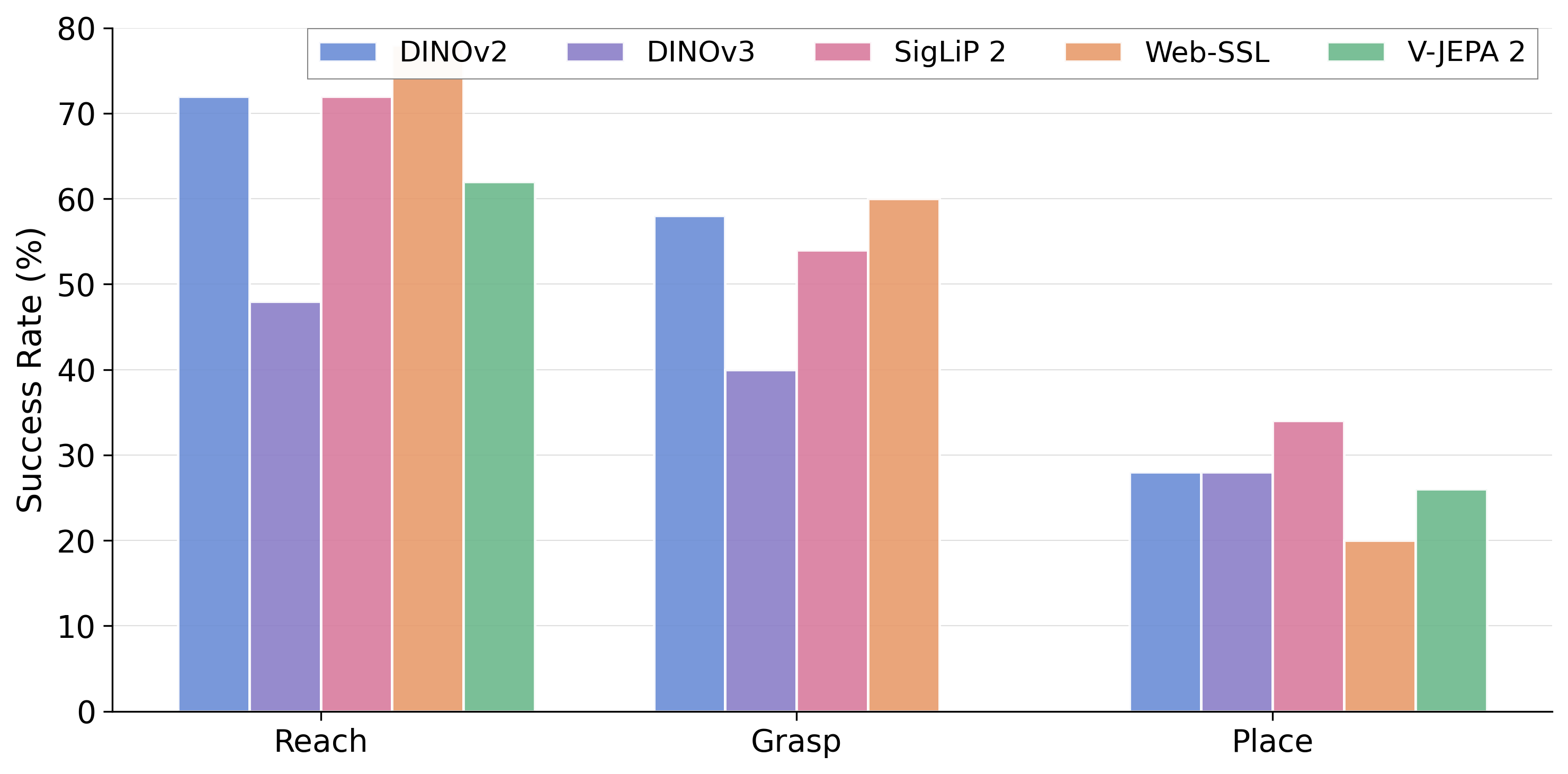}
    \caption{\textbf{Encoder Ablation.} Comparing downstream robotic task success rates on RoboCasa simulation tasks.}
    \label{fig:backbone_ablation}
    \vspace*{-0.3cm}
\end{figure}

\vspace{1.3mm}
\noindent\textbf{Model Size.} We vary the model's predictor size from \ours-S to \ours-XL (30M to 450M parameters) while keeping the training setup, data, and vision encoder consistent. See the appendix for model architecture details. ~\Cref{fig:model_size} shows that embedding prediction error and keypoint overlap percentage consistently improve with larger model size. This suggests that higher model capacity helps facilitate better dynamics learning. We use \ours-XL by default, unless otherwise noted. 
For completeness, we include the NWM$^*$~\cite{bar2025navigation} and PEVA$^*$~\cite{peva} baselines in~\Cref{fig:model_size}, and discuss these comparisons in~\Cref{sec:wm_comp}.

\vspace{1.3mm}
\noindent\textbf{Backbone.} As the pretrained vision encoder defines the latent space of our world model, we evaluate whether \ours generalizes across different backbones, an aspect not fully explored in prior work~\cite{bar2025navigation,peva,dino-wm}. We test state-of-the-art self-supervised image encoders (DINOv2~\cite{DINOv2}, DINOv3~\cite{dinov3}, Web-SSL~\cite{webssl}), video models (V-JEPA 2~\cite{vjepa2}), and language-supervised models (SigLIP 2~\cite{siglip2}). Keeping everything else fixed, we evaluate success rates in simulation, since latent spaces from different encoders are not directly comparable due to differing scales. In addition, because encoders like V-JEPA 2 and SigLIP 2 use different embedding dimensions than that of \ours's predictor, we add learnable input/output projections to align with the predictor.
\ours performs well across backbones and is not restricted to DINOv2 (\Cref{fig:backbone_ablation}), though performance varies by task, with DINOv2 strongest overall. While this demonstrates \ours's modularity with respect to the backbone, additional tuning (e.g., embedding normalization) can be important for some backbones to further improve performance.


\vspace{1.3mm}
\noindent\textbf{Hand Consistency Loss.} As detailed in \Cref{sec:training_loss}, we use the hand consistency loss as an auxiliary objective to improve fine-grained dexterity. \Cref{tab:ablation_kp} shows that adding hand consistency loss yields up to a 34\% increase in PCK@20 at 4 seconds prediction. Beyond open-loop rollout gains, predicted keypoints also benefit robot planning (\Cref{sec:planning}).

\begin{table}[t]
\centering
\footnotesize
\setlength{\tabcolsep}{6pt} 
\begin{tabular}{lcccc}
\toprule
 & \multicolumn{2}{c}{Embedding L2 Error$\downarrow$} & \multicolumn{2}{c}{PCK@20$\uparrow$} \\
\cmidrule(lr){2-3} \cmidrule(lr){4-5}
 HC Loss& At 4s & Avg & At 4s & Avg \\
 \midrule
$\times$& 0.85&  0.61&  26&  52\\
\checkmark & \cellcolor{gray!20}\textbf{0.66}&  \cellcolor{gray!20}\textbf{0.50}&  \cellcolor{gray!20}\textbf{60}&  \cellcolor{gray!20}\textbf{69}\\
\bottomrule
\end{tabular}
\caption{\textbf{Hand Consistency (HC) Loss improves fine-grained dexterity.} Reporting embedding loss and PCK@20 on EgoDex.}
\label{tab:ablation_kp}
\vspace*{-0.5cm}
\end{table}

\begin{figure}
    \centering
    \includegraphics[width=.95\linewidth]{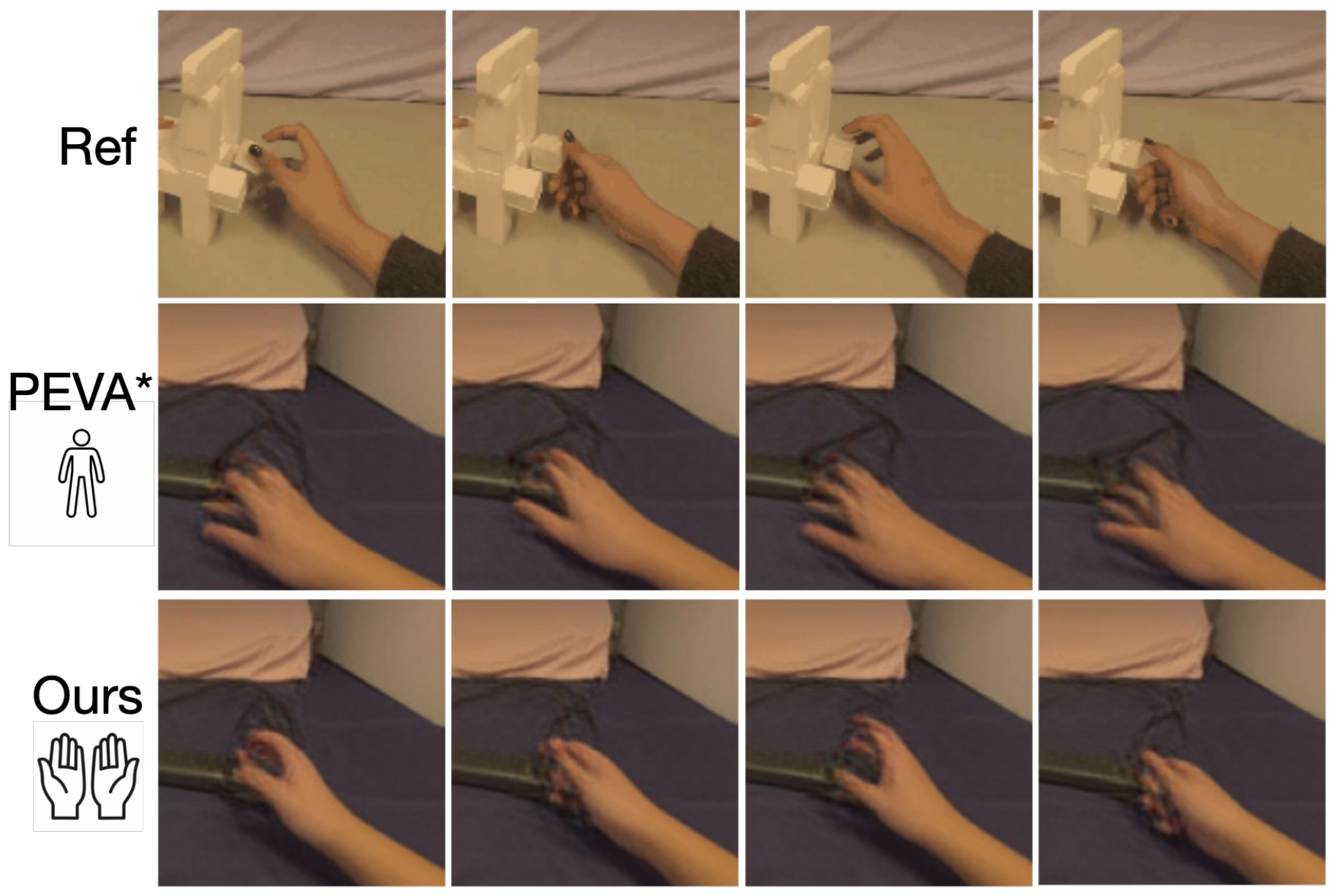}
    \caption{\textbf{Action Transfer.} Transferring actions from a reference sequence to a new environment using \ours and PEVA$^*$. \ours better captures fine-grained hands states that match those in the reference sequence.}
    \label{fig:transfer_plot}
    \vspace*{-0.2cm}
\end{figure}

\begin{figure*}[t]
    \centering
    \includegraphics[width=\linewidth]{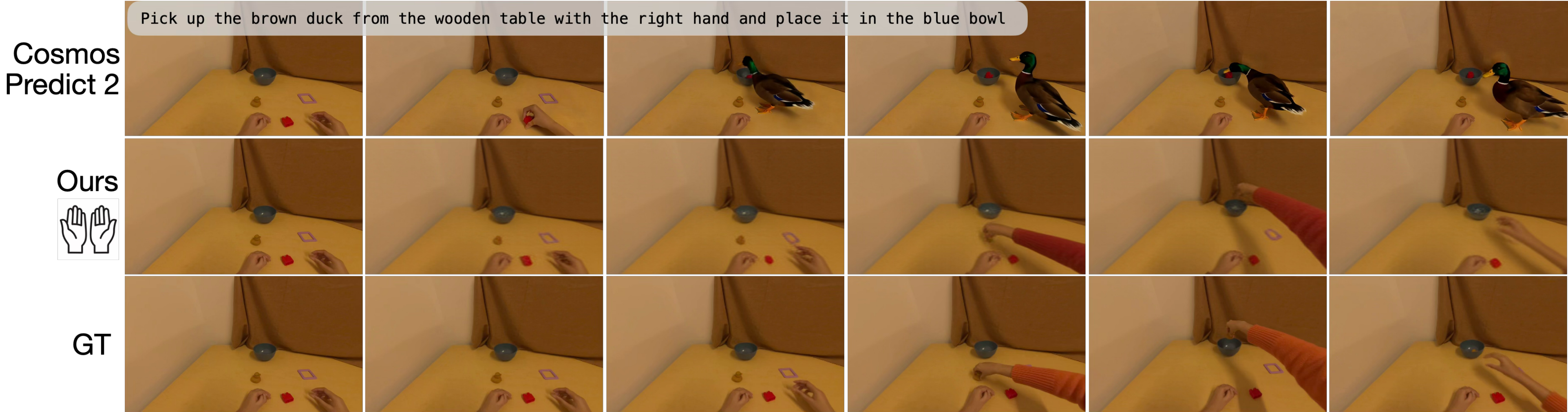}
    \caption{\textbf{Conditioning on Hand Motion Enables Precise Control}. \ours utilizes dense dexterous actions, enabling finer-grained control compared to text conditioned World Models like Cosmos-Predict 2~\citep{agarwal2025cosmos}.}
    \label{fig:cosmos}
\end{figure*}

\begin{figure*}
    \centering
    \includegraphics[width=0.95\linewidth]{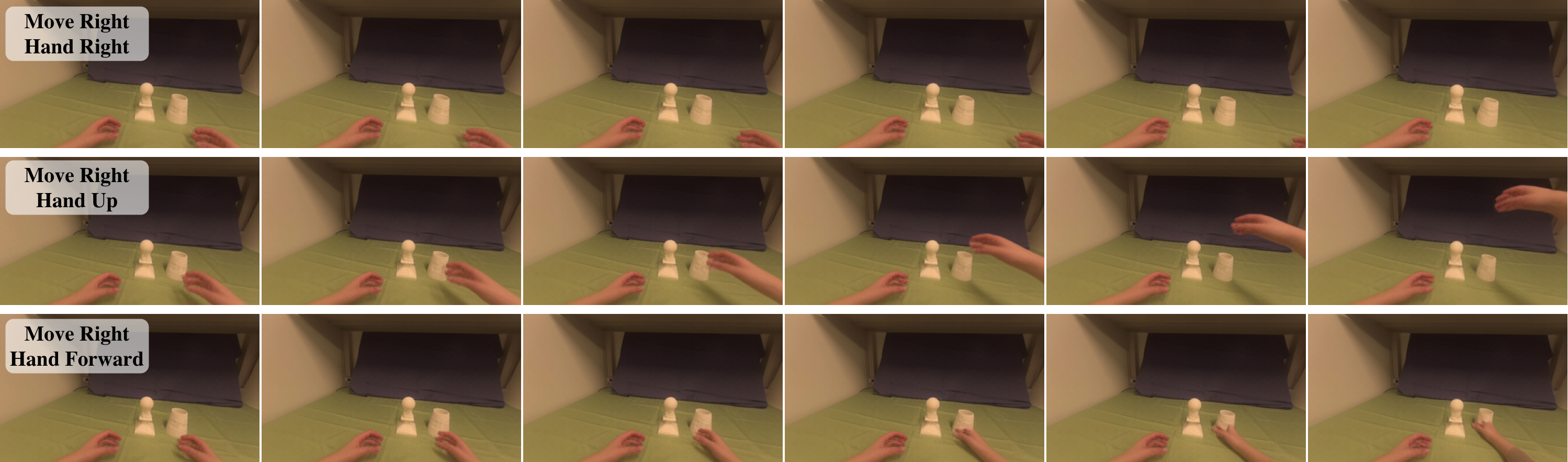}
    \caption{\textbf{Simulating Counterfactual Actions.} Starting from the same initial state, \ours predicts future states given different atomic actions. The model reliably follows each action sequence, while capturing environment dynamics (e.g, see third row final frame where the cup moves forward when the hand collides with the cup).}
    \label{fig:atomic_actions}
    \vspace*{-0.2cm}
\end{figure*}

\subsection{Baselines}
We compare \ours against several strong baselines spanning video prediction, world modeling, and action generation. For full details, see the appendix. Cosmos-Predict2~\cite{agarwal2025cosmos} is a diffusion-based “Video-to-World’’ model that synthesizes future frames from text prompts and an optional starting image. Navigation World Model (NWM)~\cite{bar2025navigation} predicts future egocentric observations conditioned on navigation actions; for fairness, we adopt a simplified variant, NWM$^*$, that conditions only on camera motion and excludes hand and body dynamics. PEVA~\cite{peva} generates egocentric videos from 3D human pose trajectories; we use a modified version, PEVA$^*$, that conditions solely on upper-body poses without finger articulation. Finally, Diffusion Policy~\cite{diff_policy} provides a state-of-the-art generative action policy that predicts multistep actions from the current observation and a goal image. 

\begin{table}[t!]
\centering
\footnotesize
\setlength{\tabcolsep}{6pt} 
\begin{tabular}{lcccc}
\toprule
Model & \multicolumn{2}{c}{Embedding L2 Error$\downarrow$} & \multicolumn{2}{c}{PCK@20$\uparrow$} \\
\cmidrule(lr){2-3} \cmidrule(lr){4-5}
 & At 4s & Avg & At 4s & Avg \\
\midrule
NWM$^*$~\cite{bar2025navigation} &   0.74 &  0.57&  34&  48\\
PEVA$^*$~\cite{peva} & \cellcolor{gray!20}\textbf{0.62}&  \cellcolor{gray!20}\textbf{0.49}&  56&  63\\
\textbf{\ours (Ours)} & 0.67&  0.51&  \cellcolor{gray!20}\textbf{60}&  \cellcolor{gray!20}\textbf{68}\\
\bottomrule
\end{tabular}
\caption{\textbf{Comparing World Models with Different Actions Spaces.} We find that \textbf{lower perceptual similarity score (Embedding L2 Error) does not always reflect more accurate hand location}, which is reliably captured by estimating the hands position, measured by percentage of correct keypoints (PCK). All models are trained on EgoDex~\cite{egodex}.}
\label{tab:wm_comparisons}
\vspace*{-0.5cm}
\end{table}

\subsection{Open-Loop Trajectory Evaluation}
\label{sec:wm_comp}

\noindent\textbf{Experiment.} We aim to test how~\ours performs on open-loop rollouts compared to challenging baselines like NWM$^*$ and PEVA$^*$. Specifically, given the initial state and an action sequence, each model predicts future latent states for $4$ seconds ($20$ frames at $5$ Hz), enabling evaluation of the world models' long-horizon prediction performance.

\vspace{1.3mm}
\noindent\textbf{Training.} We train all models on EgoDex for 40 epochs. All models use the DINOv2 encoder, allowing us to measure perceptual similarity via the same $L_2$ embedding prediction error against ground truth.

\vspace{1.3mm}
\noindent\textbf{Evaluation.} The predicted embeddings are passed through a shared keypoint prediction layer, and keypoint overlap within a $20$-pixel radius (PCK@20) is computed. Moreover, we measure the $L_2$ error, which captures overall perceptual similarity, while PCK@20 reflects local hand keypoint accuracy, important for capturing dexterous interactions.
\begin{figure*}
    \centering
    \includegraphics[width=\linewidth]{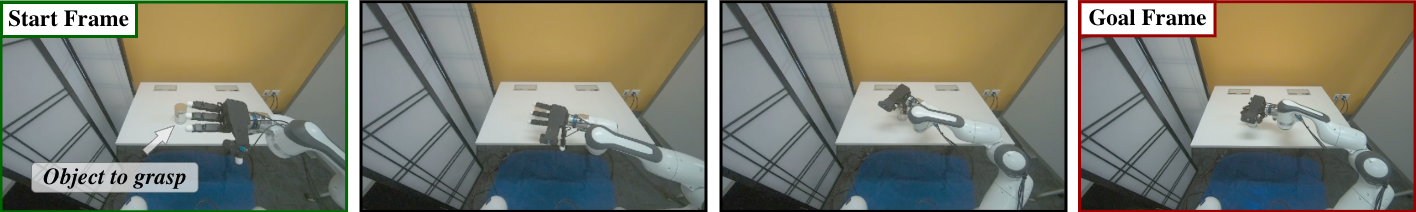}
    \caption{\textbf{Real Robot Planning Example}. Given goal and start images, \ours successfully plans and executes the trajectory by finding the optimal actions using the Cross-Entropy Method. Notably, \ours works zero-shot without any real-world training. (\Cref{sec:robot_task_exec})}
    \label{fig:real_robot_planning}
    \vspace*{-0.5cm}
\end{figure*}

\vspace{1.3mm}
\noindent\textbf{Results.} \ours outperforms other models in open-loop trajectory evaluation (\Cref{tab:wm_comparisons}). NWM$^*$ performs poorly due its sole reliance on navigation actions. PEVA$^*$ slightly outperforms \ours in $L_2$ error, however, we find that lower perceptual similarity does not always imply accurate hand position, which is important for dexterous interactions. \ours achieves over 5 points higher PCK@20 on average, indicating superior preservation of local hand information. 

\vspace{1.3mm}
\noindent\textbf{Action Transfer.} To highlight the difference from PEVA, we conduct a qualitative action transfer experiment (see~\Cref{fig:transfer_plot}). Actions from a reference trajectory are used to rollout states in a different environment. PEVA$^*$, is conditioned on body poses and produces inaccurate hand configurations, while \ours predicts fine-grained hand states that closely match those in the reference sequence.

\vspace{1.3mm}
\noindent\textbf{Comparison with Text Conditioned Models.}  While text-conditioned models like Cosmos Predict~2 generate visually compelling scenes, their predictions are not always physically grounded in hand–object interactions; for example, in~\Cref{fig:cosmos}, the model introduces a new duck simply because the text input mentions “the brown duck,” even though it refers to the dummy duck on the table.

\vspace{1.3mm}
\noindent\textbf{Controllability.} To assess the ability to follow structured physical control, we show rollouts with simple atomic actions (e.g., moving the hand up) in \Cref{fig:atomic_actions}. In each sequence, the right hand moves 1 cm per frame in the specified direction. \ours accurately follows these unseen atomic actions, and when the hand collides with the cup in the third row, the cup moves forward, indicating that \ours also learns such physical interactions.

\subsection{Human Video To Robot Transfer}
\label{sec:robot_task_exec}

\noindent\textbf{Experiment.} In addition to accurately simulating the world, \ours transfers to unseen robotic manipulation tasks in simulation and real-world when deployed on a Franka arm with Allegro grippers.
Specifically, given only small amounts of exploratory robot simulation data, our goal is to evaluate whether \ours can use the human data pretraining to perform tasks such as \textit{reach}, \textit{place}, and \textit{grasp}, which were previously unseen in the target environment and embodiment. 

\vspace{1.3mm}
\noindent\textbf{Exploratory Dataset.} 
We compare two teleoperation-free strategies for collecting exploratory data in RoboCasa~\cite{robocasa}, both using randomized environments and object placements.
\textit{Lift-Initialized Random Exploration} is the first strategy that executes ground-truth trajectories from the \textit{Lift} dataset (see appendix for details) with added noise. Environment and object randomization, and added noise, prevent successful grasps, promoting broad, unbiased coverage of the environment to bridge the embodiment gap rather than learn task-specific skills.
In the second strategy, we remove reliance on \textit{Lift} by sampling random 3D targets in randomized environments and controlling the robot to reach them. This fully programmatic data collection procedure requires no teleoperation or initialization from other datasets.
Lift-initialized Random Exploration approach achieves slightly better performance (53\% vs. 49\% average success over reach, grasp, and place simulation tasks). Therefore, in what follows, we present results corresponding to models trained on this dataset.

\vspace{1.3mm}
\noindent\textbf{Training.} We pretrain \ours on EgoDex~\cite{egodex} and DROID~\cite{droid}, then fine-tune on exploratory trajectories created in RoboCasa~\cite{robocasa}. 
We compare the fine-tuned \ours model to training a goal-conditioned Diffusion Policy (DP)~\cite{diff_policy} and a \ours model on RoboCasa from scratch, without human-data pretraining. For DP, training goals are uniformly sampled from future frames of each trajectory.

\vspace{1.3mm}
\noindent\textbf{Evaluation.} We evaluate the resulting model's planning or policy rollout performance in the real world and in simulation, without using any real-world fine-tuning data. In simulation, we conduct 50 trials each for \textit{reach}, \textit{grasp}, and \textit{place} tasks, measuring success by the Euclidean distance between target and actual poses (for \textit{reach} and \textit{place}) and additionally by object-robot contact (for \textit{grasp}). Real-world evaluation consists of 12 grasping trials with varied objects, with success determined by manually observing whether the object is in the hand. See the appendix for details.

\begin{table}[t]
\centering
\footnotesize
\setlength{\tabcolsep}{2pt} 
\begin{tabular}{lcccc}
\toprule
 & \multicolumn{3}{c}{RoboCasa} & \multicolumn{1}{c}{Real Robot} \\
\cmidrule(lr){2-4} \cmidrule(lr){5-5}
Model & Reach & Place & Grasp & Grasp \\
\midrule
Diffusion Policy~\cite{diff_policy} & 16 & 8 & 0 & 0 \\
\ours (w/o PT) & 18 & 8 & 14 & 0 \\
\textbf{\ours (Ours)} & \cellcolor{gray!20}\textbf{72} & \cellcolor{gray!20}\textbf{28} & \cellcolor{gray!20}\textbf{58} & \cellcolor{gray!20}\textbf{83} \\
\bottomrule
\end{tabular}
\caption{\textbf{Robot Transfer Results.} \ours outperforms Diffusion Policy and \ours (without Pre-Training) baselines on simulation and real robot tasks. Reporting success rates (in \%).}
\label{tab:robot_experiments}
\vspace*{-0.5cm}
\end{table}

\vspace{1.3mm}
\noindent\textbf{Simulation Transfer Results.} Planning results in~\Cref{tab:robot_experiments} show that \ours consistently outperforms all baselines. In the \textit{place} task, where the robot must maintain its grasp and transport the object without additional feedback, such as from force sensors, \ours achieves 28\% success despite the added difficulty. The large gap over the \ours variant without human video pre-training underscores the importance of human demonstrations.
Further, goal-conditioned Diffusion Policy (DP)~\cite{diff_policy} underperforms due to the exploratory nature of the training dataset, which lacks task annotations and successful completions. To verify that \ours’s gains over DP in~\Cref{tab:robot_experiments} are not due to pretraining alone, we conduct an experiment where we pre-train DP on the same human dataset (with a similar action space) and fine-tune it on robot data. DP achieves only 4\% average simulation success, far below DexWM’s 53\%, indicating that the improvement stems from modeling dexterous dynamics rather than pretraining alone. An analysis of \ours’s failure cases is in the appendix.

\vspace{1.3mm}
\noindent\textbf{Zero-Shot Real Robot Transfer Results.} We evaluate the zero-shot grasping performance of \ours when deployed on a real-world Franka arm with Allegro gripper, similar to that in the simulation~(see \Cref{tab:robot_experiments}). Without any finetuning on real robot data, \ours achieves $10$ successes out of $12$ trials ($\approx 83\%$ success rate), highlighting the planning benefits with a world model (planned trajectory example in \Cref{fig:real_robot_planning}). By planning in latent space rather than directly predicting actions, \ours shows strong generalization. In contrast, Diffusion Policy trained on exploratory simulation data fails on the real task, highlighting the robustness of world models in handling dataset quality and sim-to-real gaps.

\section{Limitations}
While we demonstrate planning with image-based goals, our framework can be extended to accommodate text-specified goals. To mitigate the embodiment gap, we rely on approximately four hours of exploratory simulation data; removing this dependency remains a direction for future work. Our current setup assumes static scenes without external agents (including pretraining datasets~\cite{egodex,droid}). Handling dynamic, interactive environments may require incorporating stochasticity, e.g., via an additional latent variable optimized at test time.

\section{Conclusion}
We explored the design of world models for learning dexterous hand–object interaction dynamics directly from human videos. A keypoint-based action representation, patch-level embeddings as state representation, and a hand-consistency loss together enable fine-grained modeling of dexterous behavior. Through several evaluations, we show that the model captures precise interaction dynamics under dexterous actions.
Moreover, we demonstrate effective transfer from human-video pretraining to previously unseen robotic tasks such as grasping, reaching, and placing. We hope that our work will inspire future exploration into dexterous modeling and world models for robotics, which will unlock generalizable robots capable of increasingly complex tasks.

\section{Acknowledgement}
The authors thank Nicolas Ballas, Naren Devarakonda, Jitendra Malik, Tushar Nagarajan, Michael Psenka, Basile Terver, and Artem Zholus for helpful discussions and feedback.

\bibliographystyle{ieeenat_fullname}
\bibliography{paper}

\clearpage
\newpage
\beginappendix

In this appendix, we describe the \ours architecture (\Cref{sec:suppl_architecture}), planning algorithm (\Cref{sec:suppl_mpc}), and training, dataset, and baseline methods (\Cref{sec:suppl_implementation_details}) in more detail. Additionally, we present further qualitative visualizations, including open-loop rollouts, counterfactual simulations, action transfer, sample executions in simulation, and sample executions in the real world (see \Cref{sec:suppl_visualization}).

\section{\ours Architecture}
\label{sec:suppl_architecture}
\subsection{Encoder}
As described in \Cref{sec:states_and_actions}, \ours leverages patch-level features extracted from the DINOv2~\cite{DINOv2} image encoder as its latent states, denoted by $s_{k_i} \in \mathbb{R}^{\mathcal{P} \times d}$. Specifically, we utilize the DINOv2-L variant with a patch size of $14$ pixels and an embedding dimension of $d=1024$. To ensure consistency across input datasets, we resize all images such that the shortest side is $224$ pixels, while preserving aspect ratio. Then, we center crop the image so that is $224 \times 392$ pixels. The encoder outputs $16 \times 28$ patches, which are then flattened to yield $\mathcal{P} = 448$ patch embeddings per image.

\subsection{Predictor}

The \ours predictor architecture is based on the Conditional Diffusion Transformer (CDiT)~\cite{bar2025navigation}, as described in \Cref{sec:predictor}. The predictor applies the CDiT block $B$ times over the input sequence of latent states, with AdaLN~\cite{peebles2023scalable} action conditioning. For all experiments in this paper, we utilize the \ours-XL variant (see Model Size ablation in \Cref{sec:ablation}). Details regarding the number of CDiT blocks, embedding dimensions, and number of attention heads for \ours-XL, \ours-L, \ours-B, and \ours-S are provided in \Cref{tab:predictor_arch_details}. For predictor variants such as \ours-B and \ours-S, where the embedding dimension differs from that of the DINOv2 encoder, we employ input and output projection layers before and after the predictor to map the latent states to the appropriate predictor and encoder dimensions, respectively.

\begin{table}[t]
\centering
\footnotesize
\setlength{\tabcolsep}{2pt} 
\begin{tabular}{lcccc}
\toprule
Model &  Params (M)&Blocks & Emb. Dim. & Heads \\
\midrule
\ours-S&  31&12& 384& 6\\
\ours-B&  104&12& 768& 12\\
 \ours-L&  344&24& 1024&16\\
\ours-XL&  456&32& 1024& 16\\
\bottomrule
\end{tabular}
\caption{\textbf{Predictor Architecture.} The number of predictor parameters (in millions), CDiT blocks, embedding dimensions, and number of attention heads are reported. Unless otherwise noted, we use \ours-XL as the default predictor for all experiments.}
\label{tab:predictor_arch_details}
\end{table}

\subsection{Keypoint Predictor}
To predict keypoints from latent states, we employ a Transformer-based keypoint prediction head. The latent states $s_{k_i} \in \mathbb{R}^{\mathcal{P} \times d}$, with shape $(16 \times 28, 1024) = (448, 1024)$, are first projected to a dimension of $256$. Learnable positional embeddings are added, and the resulting sequence is processed by $6$ Transformer blocks with $16$ attention heads each. The output is normalized and passed through a linear layer to produce a tensor of shape $(16 \times 28, 14 \times 14 \times 12)$, where $14$ denotes the patch size and $12$ is the number of keypoints to predict. This tensor is then reshaped to $(12, 16 \times 14, 28 \times 14) = (12, 224, 392)$, yielding one heatmap per keypoint. As described in \Cref{sec:training_loss}, for the HC loss, we use keypoints only on the fingertips and wrists, resulting in $12$ heatmaps: $10$ for the fingertips and $2$ for the wrists. Ground truth heatmaps are generated using unnormalized Gaussian probability density functions centered at the ground truth keypoint, with a standard deviation of 2 pixels.

\subsection{Decoder}
To visualize the latent predictions in pixel space, we train a corresponding decoder for the DINOv2 encoder following the RAE~\cite{rae} recipe. Specifically, we train a ViT-L decoder using L1, LPIPS~\cite{lpips}, and adversarial losses~\cite{goodfellow2014generative} on 50M frames randomly sampled from EgoDex~\cite{egodex} and our exploratory data collected from RoboCasa~\cite{robocasa}. The decoder reconstructs at the same resolution \ours is trained with, i.e. $224 \times 392$.

\section{Planning Optimization}
\label{sec:suppl_mpc}
In this section, we detail the planning optimization algorithm. As described in \Cref{sec:planning}, we adopt a goal-conditioned planning setup using the learned world model $f_\theta$. Given an initial state $s_0$ and a goal state $s_g$, we solve:
\begin{align}
    &\Theta_0^*, \dots, \Theta_{T-1}^* = \arg\min_{\Theta_0, \dots, \Theta_{T-1}} \; \mathcal{C}(s_T, s_g) \\
    &\text{s.t.} \quad a_k = \mathcal{G}(\Theta_k) \nonumber\\
    &\quad \quad \hat{s}_{k+1} = f_\theta(s_k, \dots, s_0, a_k), \quad k = 0, \dots, T-1
    \nonumber
\end{align}
where $\Theta_k \in \mathbb{R}^{23}$ represents the joint angles of the robot, $\mathcal{G}$ computes the action $a_k$ using forward kinematics, $\mathcal{C}$ is the planning cost, and $T$ is the planning horizon. The robot comprises $23$ joints: $7$ on the Franka Panda arm and $4$ on each of the $4$ fingers of the Allegro gripper. We assume uniformly sampled timesteps with states $\{s_0, s_1, \dots\}$ and actions $\{a_0, a_1, \dots\}$.

We employ the Cross-Entropy Method (CEM)~\cite{crossentropymethod} to optimize the joint angles $\mathbf{\Theta} = (\Theta_0, \dots, \Theta_{T-1})$. At each iteration, $N$ candidate joint angle sequences are sampled from a Gaussian distribution, mapped to actions via $\mathcal{G}$, rolled out through the world model, and evaluated using the planning cost $\mathcal{C}(s_T, s_g)$. The top-$K$ elite sequences are used to update the sampling distribution, focusing on promising trajectories. After $L$ refinement steps, we select the best joint sequence and execute the first set of joints on the robot using a low-level controller, then replan for the remaining $T-1$ steps in a receding horizon model-predictive control (MPC) fashion. Notably, during optimization, the camera motion components of the action vector are excluded, as the camera remains stationary.

\begin{figure}[t]
    \centering
    \includegraphics[width=\linewidth, trim=0 0 0 0, clip]{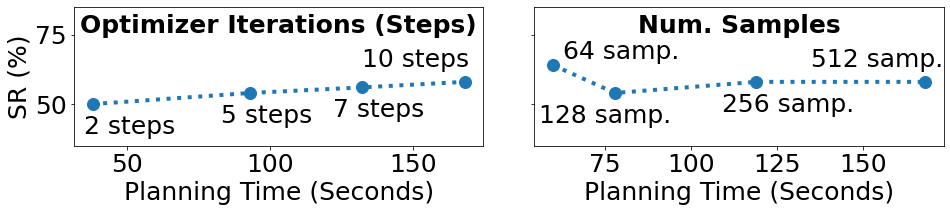}
    \caption{Planning time and success rate as optimizer iterations (with samples=512) and sample count (with optimizer iterations=10) vary.}
    \label{fig:planning_time}
\end{figure}

For simulation tasks (Section 4.5), we use $\{T, N, K, L\} = \{3, 512, 10, 10\}$, while for real robot experiments, we use $\{2, 256, 10, 10\}$. In real-world experiments, we do not employ receding horizon planning and execute the two-step optimized actions in open loop. The low-level controllers in both simulation and real robot settings take the optimized joint angles as targets, interpolate a trajectory between the current and target joints, and execute this on the robot with small control steps. \ours rollouts during simulation planning are parallelized across 8 H100 GPUs. 
~\Cref{fig:planning_time} shows the trade-off between planning time per episode and success rate as the sample count and iterations vary on the grasping task. 
Across all tested settings, performance is similar while planning time ranges from 38 to 168 seconds (default is 168 sec. for maximum robustness; lower settings can be used for reduced compute).

We find that good initialization of the Gaussian distribution parameters (mean and variance) is crucial for optimization performance. In our experiments, the mean is initialized to the robot joint configuration of the starting state. The standard deviations for the Franka arm joints are set to $0.3$ to enable broad exploration, while those for the Allegro gripper are set to $0.1$ for finer control ($0.01$ in the \textit{place} task to aid gripping). For the \textit{grasp} task, we teleoperate the robot to perform a `dummy' \textit{grasp} motion (approaching the countertop and closing the gripper), and uniformly sample $T$ steps from this sequence to initialize the mean for the \textit{grasp} task.

\section{Implementation Details}
\label{sec:suppl_implementation_details}
\subsection{Training Details}
As detailed in \Cref{sec:training_loss} we keep the pre-trained DINOv2 encoder frozen during training and optimize the rest of the model using the loss $\mathcal{L} = \mathcal{L}_{state} + \lambda \mathcal{L}_{HC}$, with $\lambda = 100$ yielding the best results. The decoder is trained separately, as it is used only for visualization purposes. The model is trained on the EgoDex~\cite{egodex} and DROID~\cite{droid} datasets with a batch size of $4096$, using the Adam~\cite{kingma2014adam} optimizer with an initial learning rate of $10^{-4}$, which is reduced to $10^{-7}$ over $40$ epochs via a cosine annealing schedule~\cite{smith2019super}.

During training, we randomly select a frame $s_{k_{n+1}}$ from the current video sequence and choose $8$ preceding frames at non-uniform intervals, following the strategy of PEVA~\cite{peva}, with a maximum window size of 4 seconds. This forms the states $\{s_{k_0}, s_{k_1}, \dots, s_{k_n}\}$ as illustrated in Figure 2. To provide rich training signals, the predictor is tasked with predicting $s_{k_{i+1}}$ using the context $\{s_{k_0}, \dots, s_{k_i}\}$ for $i = \{1, \dots, 9\}$. Since the videos in the datasets can be longer than 4 seconds, we ensure uniform sampling by dividing each video into $10$ equal segments and randomly selecting $s_{k_{n+1}}$ from each segment in different data loading iterations.

The model is subsequently fine-tuned on the exploratory RoboCasa simulation dataset for $50$ epochs with a batch size of $8$, using the Adam~\cite{kingma2014adam} optimizer and a learning rate of $10^{-5}$. Similar to the training on EgoDex and DROID, we randomly select $\{s_{k_0}, s_{k_1}, \dots, s_{k_n}, s_{k_{n+1}}\}$ from the sequences for training. Notably, incorporating multistep prediction during fine-tuning (see \Cref{sec:predictor}) improves performance. Consequently, we adopt this approach in our experiments.

\subsection{Datasets and Robotic Benchmarks}
\label{sec:supp_dataset}
For training, we use EgoDex~\cite{egodex}, an egocentric human dataset, and DROID~\cite{droid}, a robotics dataset. Additionally, we fine-tune the model for robotic tasks using exploratory data collected with the RoboCasa~\cite{robocasa} simulator, as described below.

\vspace{1.3mm}
\noindent\textbf{EgoDex~\cite{egodex}.} An egocentric video dataset for learning dexterous manipulation, recorded using Apple Vision Pro headsets. The dataset comprises $829$ hours of $1080$p egocentric video at $30$ Hz, containing $194$ manipulation tasks involving $500$ distinct objects. EgoDex provides rich multimodal annotations, including 3D skeletal poses for the upper body and $25$ keypoints per hand, camera intrinsics and extrinsics, and confidence scores for all pose estimates. Approximately 1\% of the data has been set aside as test data, following the original split.

\vspace{1.3mm}
\noindent\textbf{DROID \cite{droid}.} A diverse robot manipulation dataset collected using the Franka Panda robot equipped with parallel-jaw grippers, with data captured from multiple camera viewpoints. To approximate dexterous hand movements, dummy hand keypoints, selected on concentric circles centered at the end-effector, are generated based on the gripper's open/close status (\Cref{fig:droid_hands}). We use about 100 hours of DROID data for training \ours, in conjunction with EgoDex.

\begin{figure*}[t!]
    \centering
    \includegraphics[width=\linewidth]{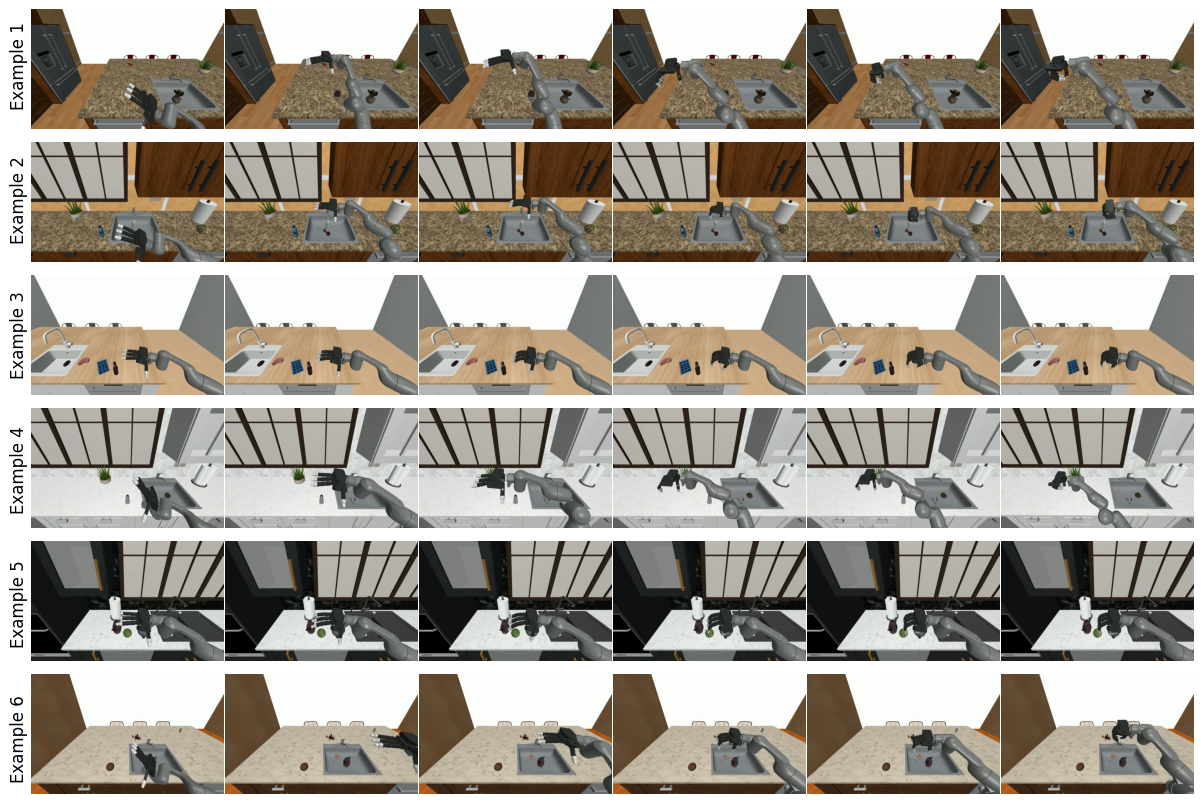 }
    \caption{\textbf{Exploratory Data Examples.} Sequences of the robot performing random movements in the environment were collected to fine-tune \ours for application to robotic simulation and real-world tasks. This exploratory data provides broad, unbiased coverage of the environment, aiming to bridge the embodiment gap between human and robot.}
    \label{fig:exploratory_data}
\end{figure*}

\vspace{1.3mm}
\noindent\textbf{RoboCasa~\cite{robocasa}.} As described in \Cref{sec:robot_task_exec} of the manuscript, we evaluate \ours on robotic simulation tasks using the RoboCasa simulation framework~\cite{robocasa} with the Franka TMR platform (two Franka Panda arms equipped with Allegro grippers). In this work, only the right arm and gripper are used; the remaining components remain static. RoboCasa features 120 photorealistic kitchen scenes spanning 10 diverse floor plans and 12 architectural styles, with textures procedurally generated using generative AI tools to maximize visual diversity. The simulator supports multiple robot embodiments, including mobile manipulators, humanoids, and quadrupeds, facilitating cross-platform policy learning.

We evaluate our method on \textit{reach}, \textit{grasp}, and \textit{place} tasks within RoboCasa. In these tasks, the robot must reach a location, grasp an object, or place an object at a goal specified by an image, respectively. For each test task, we evaluate 50 trajectories with randomized environments and object placements.
To bridge the gap between human and robot embodiments, we fine-tune \ours on approximately 4 hours of exploratory sequences of random arm movements collected in RoboCasa. As described in \Cref{sec:robot_task_exec}, this exploratory data is generated by initializing environments with sequences from the Lift dataset (detailed below), randomly permuting objects to prevent successful grasps, and replaying Lift actions with added continuous noise. This approach ensures broad, unbiased coverage of the environment, aiming to bridge the embodiment gap rather than to learn task-specific skills. Additionally, the dataset includes sequences where the robot only opens and closes its gripper without arm movement. Examples of these exploratory sequences are shown in \Cref{fig:exploratory_data}.

\textbf{Lift} is a dataset created by controlling the robot to grasp objects and lift them into the air, using varied environment configurations such as backgrounds and object placements. Sequences from this dataset are also used for testing models in the RoboCasa section of \Cref{tab:dataset_ablation}. 
While exploratory data could also be generated by other means, using Lift was more convenient, less labor-intensive, and ensured the robot remained within environment bounds.

Notably, the Allegro gripper has four fingers per hand, compared to five in humans. \ours's action space is defined for five-fingered hands (see \Cref{sec:states_and_actions}). To address this, we duplicate the keypoints of the last Allegro finger (equivalent to the human ring finger) and assign these values to the pinky finger in the action vector. Since only the right robot hand and gripper are used, actions corresponding to the left hand are set to zero.

\vspace{1.3mm}
\noindent\textbf{Real-World.} We use the same robotic setup for zero-shot real-world experiments as in simulation, evaluating the grasp task over 12 trajectories with 4 different objects and varied object configurations. 


\subsection{Baselines}
\label{sec:supp_baselines}
\vspace{1.3mm}
\noindent\textbf{Cosmos-Predict2~\cite{agarwal2025cosmos}.} A diffusion-based model for video generation from a text prompt (``Video-to-World'') and an optional starting image. It predicts future frames conditioned on scene descriptions, enabling high-fidelity and temporally coherent video synthesis. 

\vspace{1.3mm}
\noindent\textbf{Navigation World Model (NWM)~\cite{bar2025navigation}.} NWM is a controllable video generation model that predicts future egocentric observations from past frames and navigation actions. For fair comparison, we implement NWM in our framework by conditioning only on navigation (camera movement) actions, excluding hand and body motion. This variant is referred to as NWM$^*$ in \Cref{tab:wm_comparisons} and \Cref{fig:model_size}.

\vspace{1.3mm}
\noindent\textbf{PEVA \cite{peva}.} A whole-body conditioned video prediction model that generates egocentric videos from 3D human pose trajectories. Similar to NWM$^*$, we implement PEVA within our framework by conditioning on the upper body human poses, excluding the fingers. This variant is referred to as PEVA$^*$ in \Cref{tab:wm_comparisons} and \Cref{fig:model_size}. 

\vspace{1.3mm}
\noindent\textbf{Diffusion Policy \cite{diff_policy}.} A state-of-the-art generative behavior cloning method that models action sequences using a denoising diffusion process. Conditioned on the current observation and an image goal, the model outputs a sequence of actions to execute in the environment. We adopt the official implementation from \cite{diff_policy}, using a transformer backbone for the policy. The policy uses a context window of 2 and predicts an action chunk of length 9. Observations are encoded using the average-pooled DINOv2 patch features. To incorporate goal conditioning, we follow \cite{cui2022playpolicyconditionalbehavior} and randomly sample a future frame from the same trajectory to serve as the goal during training.

\subsection{Success Criteria for Simulation Tasks}
\label{sec:supp_success_criteria}
\vspace{1.3mm}
\noindent \textbf{Reach:} Success is reported when the average Euclidean distance between the 3D positions of the fingertips and wrists in the frame reached by the robot and goal frame is less than 15~cm for 10 consecutive time steps (approximately 1 second).

\vspace{1.3mm}
\noindent \textbf{Grasp:} Success is reported when (a) the Euclidean distance between the robot wrist and the object to be grasped is less than 20~cm, and (b) contact is detected between the robot gripper and the object for 10 consecutive time steps (approximately 1 second).

\vspace{1.3mm}
\noindent \textbf{Place:} Success is reported when the distance between the final position of the manipulated object and its position in the goal frame is less than 10~cm for 10 consecutive time steps (approximately 1 second).

\subsection{Success Criteria for Real-World Grasping Tasks}
Since it is not trivial to automatically find the distance between objects or detect contact in the real world, success is determined by manual observation of whether the object is securely held in the gripper.

\subsection{Manipulation Failure Cases Breakdown}
\label{sec:failure_breakdown}
\Cref{tab:failure_breakdown} categorizes failures based on where the manipulation pipeline breaks down. Contact failures arise from complex contact dynamics during interaction with the object. Drop refers to cases where the robot drops the object before reaching the target location. Last cm errors occur when the robot reaches the target region but fails due to small inaccuracies in the final few centimeters. State Drift denotes failures where the world model’s state prediction gradually drifts over time.

\begin{table}[h]
\centering
\setlength{\tabcolsep}{2pt}
\begin{tabular}{l |c |c |c |c |c}
\hline
\textbf{Task} & \textbf{Contact} & \textbf{Drop} & \textbf{Last cm} & \textbf{State Drift} & \textbf{Total} \\
\hline
\textbf{Reach} & 0 & 0  & 24 & 4 & 28\\
\textbf{Grasp} & 4 & 0  & 20 & 18 & 42\\
\textbf{Place} & 0 & 70 & 0  & 2 & 72\\
\hline
\end{tabular}
\caption{Failure Breakdown (\%) into \textbf{Contact} Dynamics, Object \textbf{Drop}ped before Target, \textbf{Last cm} error, and Latent\textbf{ State Drift}.}
\label{tab:failure_breakdown}
\end{table}

\section{Additional Visualizations}
\label{sec:suppl_visualization}
\begin{figure*}
    \centering
    \includegraphics[width=\linewidth]{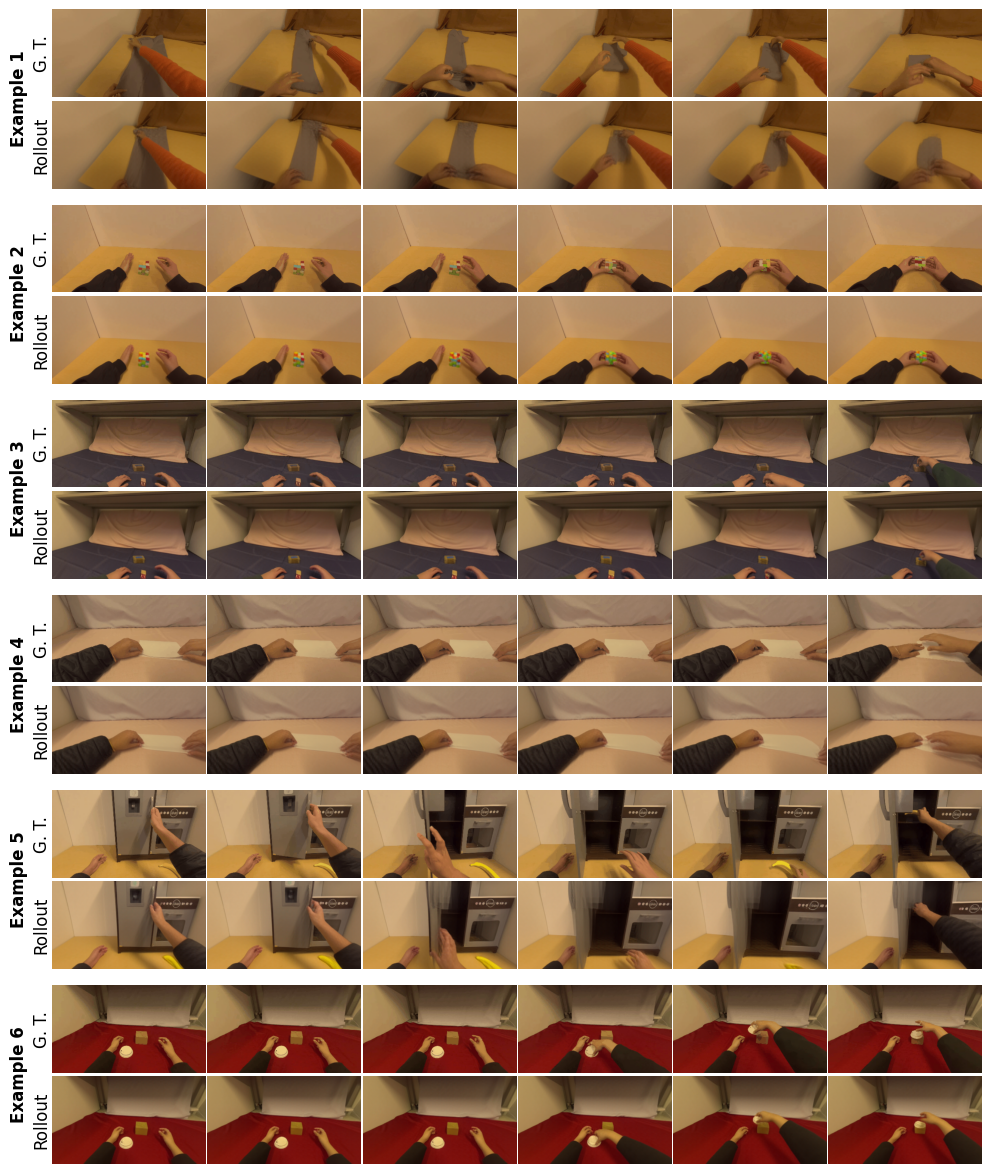}
    \caption{\textbf{Open-Loop Trajectory Rollouts.} Given the initial state and an action sequence, \ours predicts future latent states over a 4-second horizon, rolling out 20 frames at 5~Hz. For visualization, predicted frames are subsampled in the figure due to space constraints. Latent states are decoded into images for visualization. The predicted rollout closely follows the ground truth trajectory even in long-horizon complex dexterous tasks.}
    \label{fig:supp_rollouts}
\end{figure*}

\begin{figure*}
    \centering
    \includegraphics[width=\linewidth]{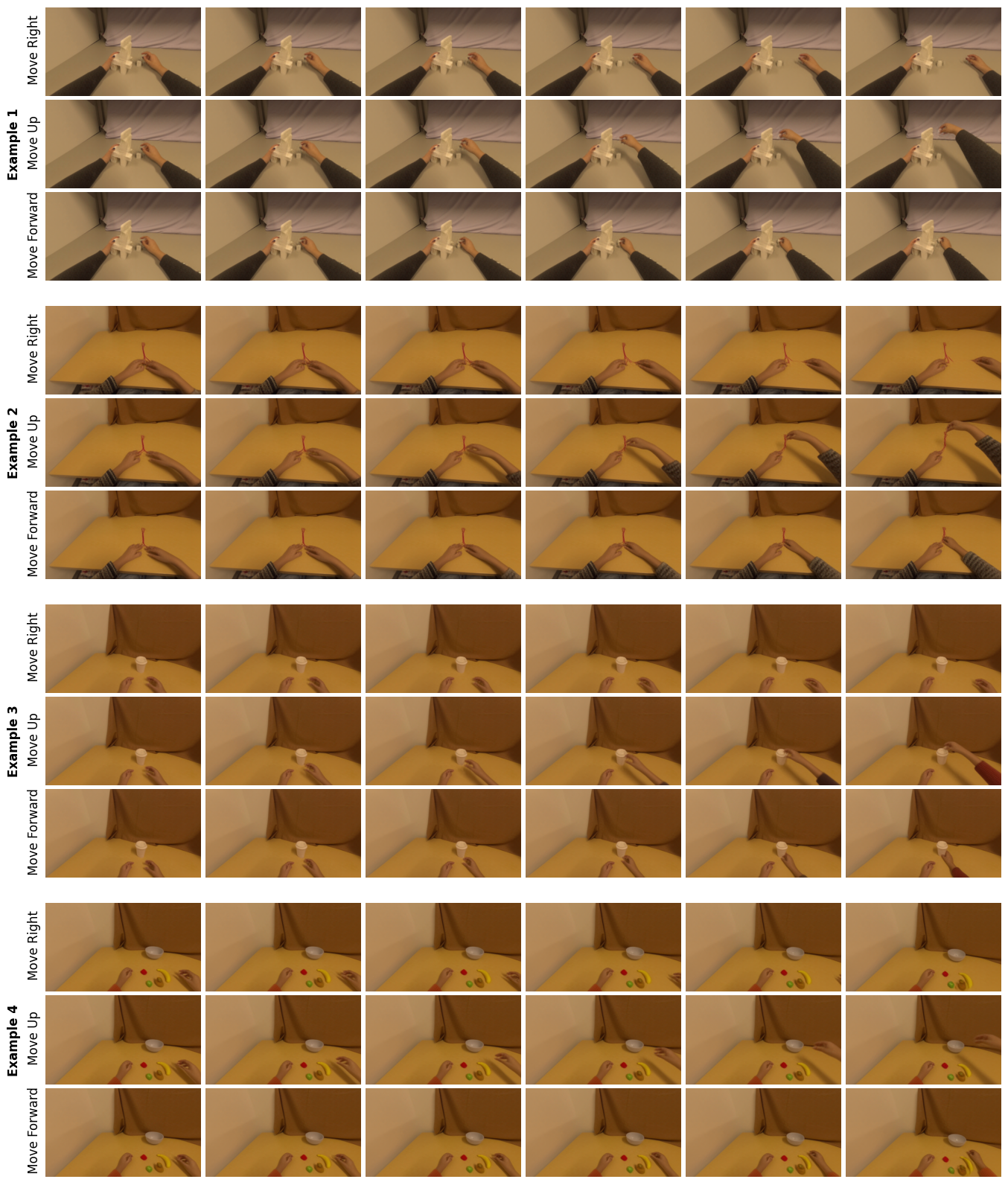 }
    \caption{\textbf{Simulating Counterfactual Actions.} Starting from the same initial state, \ours predicts future states given different atomic actions for controlling the right hand. The model reliably follows each action sequence while accurately capturing environment dynamics (e.g., pulling the string upward in \textit{Move Up} in Example~2, and pushing the cup forward in \textit{Move Forward} in Example~3.)}
    \label{fig:supp_atomic_actions}
\end{figure*}

\begin{figure*}[t!]
    \centering
    \includegraphics[width=\linewidth]{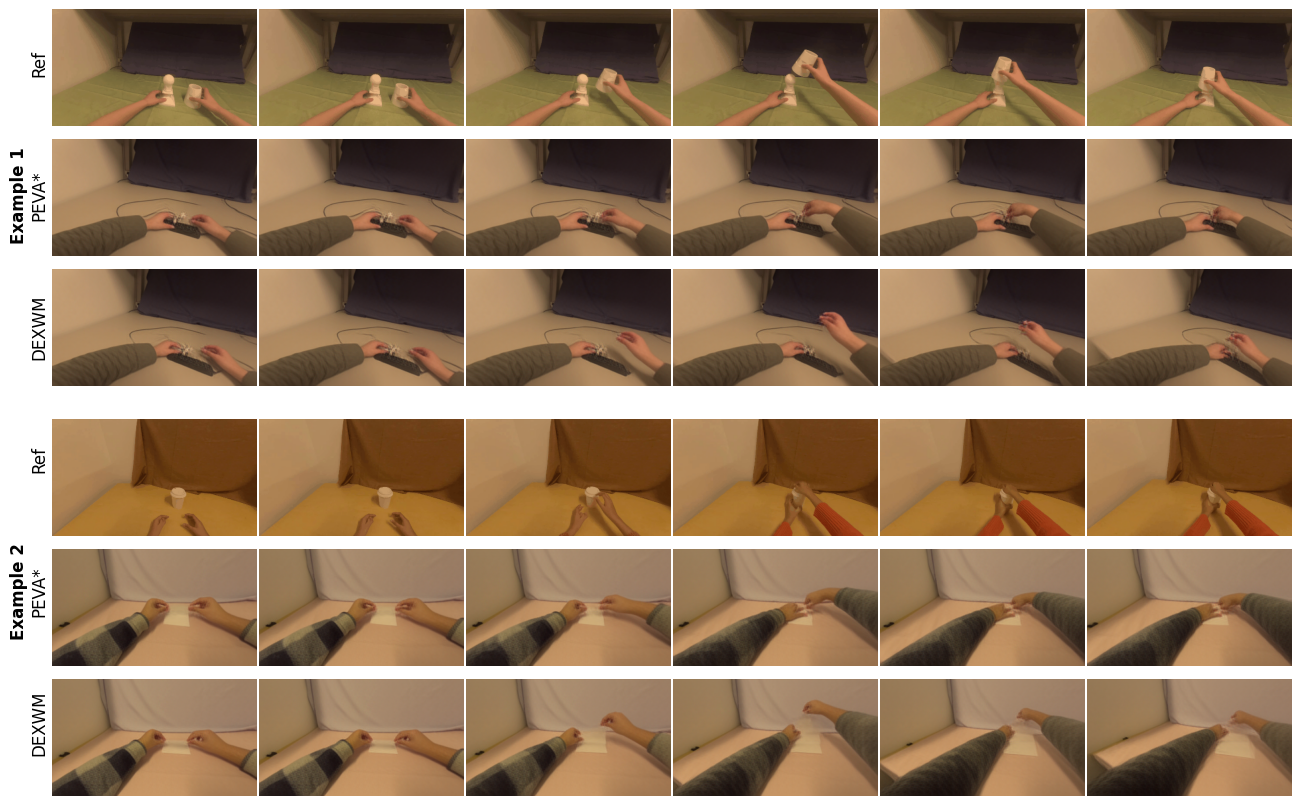 }
    \caption{\textbf{Action Transfer.} Transferring actions from a reference sequence to a new environment using \ours and PEVA$^*$. \ours better captures fine-grained hands states that match those in the reference sequence.}
    \label{fig:action_transfer}
\end{figure*}

\begin{figure*}[t!]
    \centering
    \includegraphics[width=\linewidth]{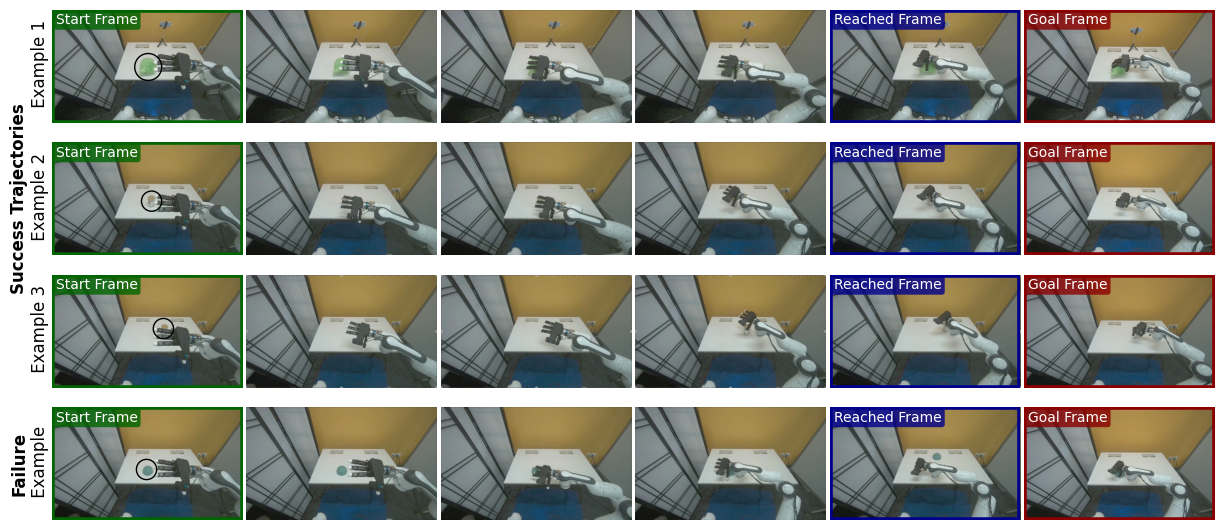 }
    \caption{\textbf{Real Robot Task Examples.} The robot successfully grasps objects, which are highlighted by black circles in the first image of each sequence for visual clarity, given the goal (red) and start (green) images. Notably, \ours operates in a zero-shot manner, without any real-world training. In some cases, such as examples 2 and 3, the object moves slightly from its original location after being actually grasped by the gripper. We also present a failure trajectory, where a collision between the grippers and an upside-down bowl causes the bowl to move away, resulting in a missed grasp.}
    \label{fig:real_world_experiments}
\end{figure*}

\begin{figure*}[t!]
    \centering
    \includegraphics[width=\linewidth]{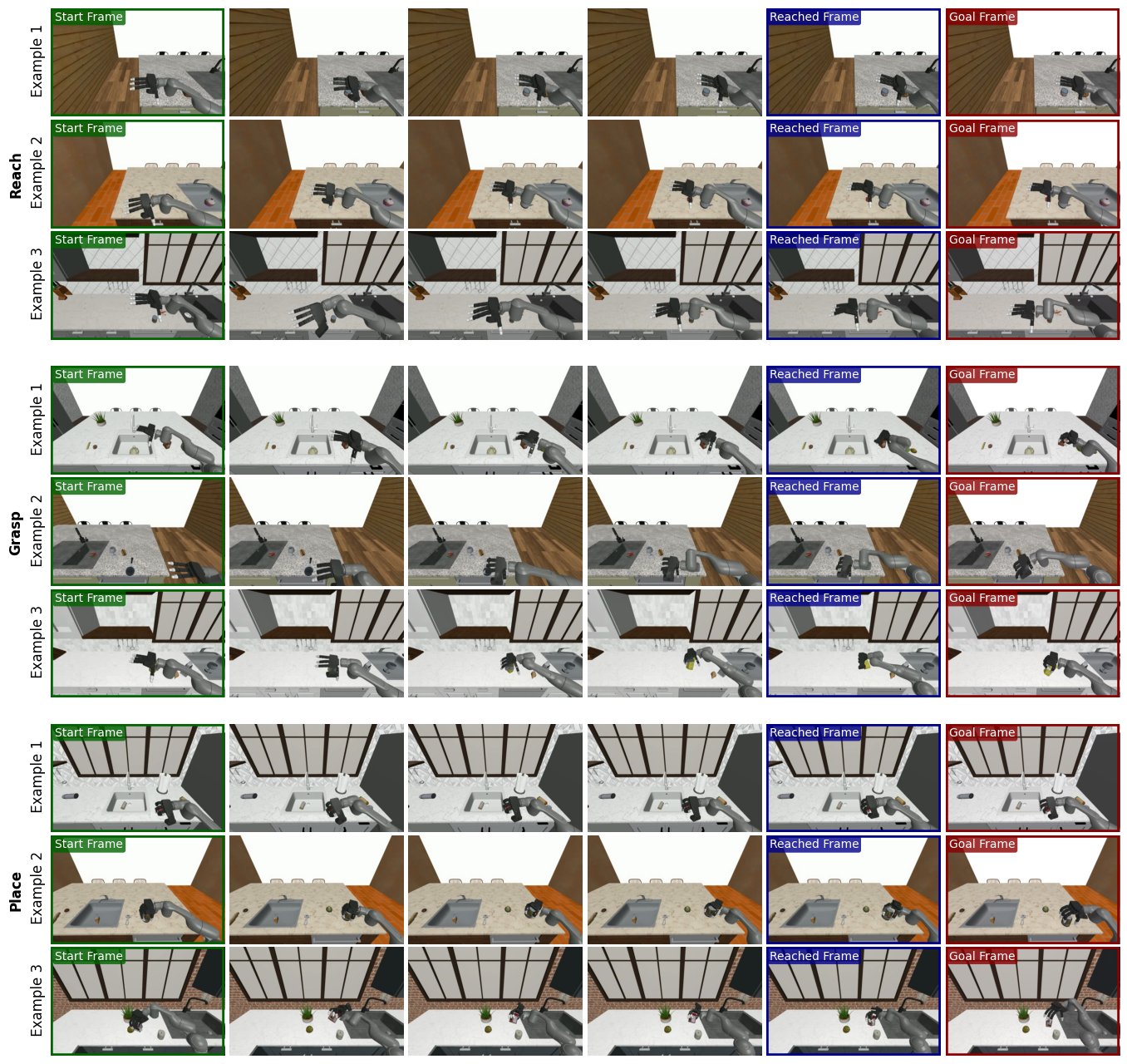 }
    \caption{\textbf{Robot Task Example in Simulation.} Given goal (red) and start (green) images, \ours successfully plans the trajectory by finding optimal actions using the Cross-Entropy Method inside an MPC framework. The final reached state (blue) in each task closely resembles the goal frame, demonstrating successful execution. Notably, \ours was not trained on any task-specific robot data.}
    \label{fig:sim_experiments}
\end{figure*}

In the following, we show additional visualizations of \ours in open-loop trajectory rollout (\Cref{fig:supp_rollouts}), simulating counterfactual actions (\Cref{fig:supp_atomic_actions}), action transfer from reference sequences (\Cref{fig:action_transfer}), real world tasks (\Cref{fig:real_world_experiments}), and example executions from robotic simulation (\Cref{fig:sim_experiments}).


\end{document}